\documentclass[10pt,twocolumn,letterpaper]{article}

\usepackage{cvpr}              %
\usepackage{tabularx}
\usepackage{multirow}
\usepackage{soul}
\usepackage{marvosym} %

\usepackage[dvipsnames]{xcolor}

\newcommand{\bK}{\mathbf{K}}

\newcommand{\bx}{\mathbf{x}}

\newcommand{\bI}{\mathbf{I}}

\newcommand{\bS}{\mathbf{S}}
\newcommand{\bs}{\mathbf{s}}

\newcommand{\bW}{\mathbf{W}}

\newcommand{\bR}{\mathbf{R}}

\newcommand{\bt}{\mathbf{t}}
\newcommand{\bJ}{\mathbf{J}}

\newcommand{\bD}{\mathbf{D}}

\newcommand{\bC}{\mathbf{C}}
\newcommand{\bA}{\mathbf{A}}

\newcommand{\bff}{\mathbf{f}}
\newcommand{\bF}{\mathbf{F}}

\newcommand{\bc}{\mathbf{c}}

\newcommand{\bb}{\mathbf{b}}

\newcommand{\bmu}{\boldsymbol{\mu}}
\newcommand{\bSigma}{\boldsymbol{\Sigma}}

\newcommand{\nR}{\mathbb{R}}

\newcommand{\cN}{\mathcal{N}}

\newcommand{\cL}{\mathcal{L}}

\makeatletter
\DeclareRobustCommand\onedot{\futurelet\@let@token\@onedot}
\def\@onedot{\ifx\@let@token.\else.\null\fi\xspace}

\def\wrt{wrt\onedot}

\def\Fig{Fig\onedot}   
\makeatother

\newcommand{\figref}[1]{\Fig~\ref{#1}}
\newcommand{\secref}[1]{Section~\ref{#1}}

\renewcommand{\eqref}[1]{Eq.~\ref{#1}}
\newcommand{\tabref}[1]{Table~\ref{#1}}

\newcommand{\boldparagraph}[1]{\vspace{0.2cm}\noindent{\bf #1:} }

\definecolor{orange}{rgb}{1,0.5,0}

\newif\ifcomment
\commenttrue
\ifcomment
    \newcommand{\ag}[1]{ \noindent {\color{orange} {\bf Andreas:} {#1}} }
    \newcommand{\yl}[1]{ \noindent {\color{cyan} {\bf Yiyi:} {#1}} }

\else
	\newcommand{\ag}[1]{}
	\newcommand{\yl}[1]{}
\fi

\newcolumntype{P}[1]{>{\centering\arraybackslash}m{#1}}

\definecolor{cvprblue}{rgb}{0.21,0.49,0.74}
\usepackage[pagebackref,breaklinks,colorlinks,citecolor=cvprblue]{hyperref}

\def\paperID{****} %
\def\confName{CVPR}
\def\confYear{2024}

\makeatletter
\def\blfootnote{\gdef\@thefnmark{}\@footnotetext}
\makeatother

\title{HUGS: Holistic Urban 3D Scene Understanding via Gaussian Splatting}

\author{
Hongyu Zhou$^{1}$,
Jiahao Shao$^{1}$,
Lu Xu$^{1}$,
Dongfeng Bai$^{2}$,
Weichao Qiu$^{2}$,
Bingbing Liu$^{2}$
\\
Yue Wang$^{1}$,
Andreas Geiger$^{3,4}$,
Yiyi Liao$^{1}\textsuperscript{\Letter}$
\vspace{0.2cm}
\\
{\normalsize $^{1}$ Zhejiang University \quad $^{2}$ Huawei Noah's Ark Lab}
{\normalsize \quad $^{3}$ University of Tübingen \quad }
{\normalsize $^{4}$ Tübingen AI Center}
\\
}

\begin{document}

\twocolumn[{%
\renewcommand\twocolumn[1][]{#1}%
\maketitle
\vspace{-1cm}
\begin{center}
    \centering
    \captionsetup{type=figure}
    \includegraphics[width=\textwidth]{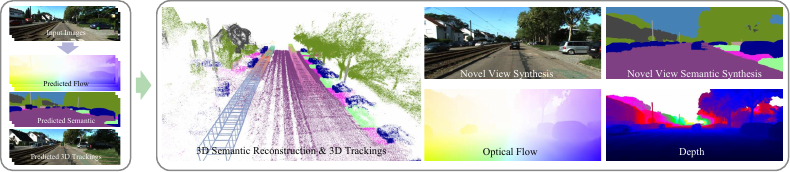}
    \captionof{figure}{\textbf{Illustration.} Given posed RGB images as input, our method lifts noisy 2D \& 3D predictions to the 3D space via decomposed 3D Gaussians, and enables holistic scene understanding in 2D and 3D space.  }
    \label{fig:teaser}
\end{center}%
}]

\blfootnote{$\textsuperscript{\Letter}$Corresponding author.}

\begin{abstract}
Holistic understanding of urban scenes based on RGB images is a challenging yet important problem. It encompasses understanding both the geometry and appearance to enable novel view synthesis, parsing semantic labels, and tracking moving objects.
Despite considerable progress, existing approaches often focus on specific aspects of this task and require additional inputs such as LiDAR scans or manually annotated 3D bounding boxes. In this paper, we introduce a novel pipeline that utilizes 3D Gaussian Splatting for holistic urban scene understanding.
Our main idea involves the joint optimization of geometry, appearance, semantics, and motion using a combination of static and dynamic 3D Gaussians, where moving object poses are regularized via physical constraints.
Our approach offers the ability to render new viewpoints in real-time, yielding 2D and 3D semantic information with high accuracy, and reconstruct dynamic scenes, even in scenarios where 3D bounding box detection are highly noisy. 
Experimental results on KITTI, KITTI-360, and Virtual KITTI 2 demonstrate the effectiveness of our approach. Our project page is at 
\href{https://xdimlab.github.io/hugs_website}{https://xdimlab.github.io/hugs\_website}.
\end{abstract}

\vspace{-0.5cm}
\section{Introduction}

Reconstructing urban scenes is an important task in computer vision with numerous applications. Consider the creation of a photorealistic simulator for autonomous driving, in this context, it becomes crucial to holistically represent all aspects of the scene relevant to driving. This entails tasks like synthesizing images at interpolated and extrapolated viewpoints in real-time, reconstructing 2D and 3D semantics, generating depth information, and tracking dynamic objects. 
To minimize sensor cost, achieving such a holistic understanding exclusively from posed RGB images holds significant value.

With the rise of neural rendering, many approaches have emerged to lift 2D information to 3D space, enabling scene understanding based solely on RGB images.
Several previous works focus on reconstructing static urban scenes,  achieving high-quality novel view appearance and semantic synthesis \cite{rematas_urban_2021, fu_panoptic_2022, zhang_nerflets_2023}. Another line of work addresses dynamic scenes \cite{ost_neural_2021, kundu_panoptic_2022, yang_unisim_nodate, wu_mars_2023}, but most of them require ground truth 3D bounding boxes of dynamic objects as input, which are costly to acquire. PNF~\cite{kundu_panoptic_2022} is the only method that utilizes noisy bounding boxes obtained through monocular 3D detection and tracking, where the transformations of the bounding boxes are jointly optimized during training. However, na\"ive joint optimization of per-frame pose transformations is prone to local minima and sensitive to the initialization. Furthermore, while existing methods are capable of rendering accurate 2D semantic labels, it is non-trivial to extract accurate semantics in 3D due to the inaccurate (inferred) 3D geometry.
In addition, most of these methods are unable to achieve real-time rendering.

In this paper, We leverage predicted 2D semantic labels, optical flow, and 3D tracks, despite their inherent noise and imperfections, to achieve a holistic understanding of the dynamic scenes based on RGB images (see \figref{fig:teaser}). Towards this goal, we infer geometry, appearance, semantics, and motion in 3D space using a decomposed scene representation. We leverage 3D Gaussians as the scene representation, which have recently demonstrated superior novel view synthesis performance on static scenes with real-time rendering capability~\cite{kerbl_3d_nodate}. Specifically, we propose to decompose the scene into static regions and rigidly moving dynamic objects. We model the poses of these moving objects while adhering to the physical constraints of a unicycle model, effectively reducing the impact of noise during tracking and leading to superior performance compared to optimizing object poses individually. This allows us to reconstruct dynamic scenes even when 3D bounding box predictions are highly noisy. 
Further, we extend 3D Gaussian Splatting to model camera exposure and explore initialization on dynamic scenes, enabling state-of-the-art novel view synthesis performance on urban scenes. Additionally, we incorporate semantic information into 3D Gaussians, enabling the rendering of semantic maps and the extracting of 3D semantic point clouds. Finally, we integrate the RGB, semantics and optical flow to jointly supervise the model training, and investigate the interaction between these image cues to improve the performance of the scene understanding tasks.  %

Our main contributions are as follows: 
1) Our method addresses the task of dynamic 3D urban scene understanding by extending Gaussian Splatting to model additional modalities, including semantic, flow, and camera exposure, as well as dynamic objects.
2) We achieve the decomposition of static and multiple dynamic objects from sparse urban images and noisy labels by incorporating physical constraints, omitting the requirement of ground truth 3D bounding boxes for reconstructing dynamic scenes.
3) Our method achieves state-of-the-art performance on various benchmarks, including novel view appearance and semantic synthesis, as well as 3D semantic reconstruction.

\section{Related Work}

\boldparagraph{3D Scene Understanding}
Understanding urban scenes from various aspects has been considered essential for autonomous driving. Numerous techniques have focused on predicting semantic labels \cite{tao_hierarchical_2020, borse_inverseform_2021, cheng_panoptic-deeplab_2020}, depth maps \cite{piccinelli_idisc_2023, eftekhar_omnidata_2021}, and optical flows \cite{xu_unifying_2023-1} solely from 2D input images. While these methods have demonstrated impressive accuracy within the confines of the 2D space, they often fall short of grasping a profound understanding of the underlying 3D environment. Consequently, this limitation can hinder the multi-view consistency of their predictions.
Another line of approach suggests conducting semantic scene understanding solely based on 3D input \cite{qi_pointnet_2017, robert_learning_2022}. This approach heavily relies on LiDAR input, which is known to be costly and resource-intensive to collect.

More recently, a particular approach has emerged, aiming to elevate 2D information to the 3D space to facilitate scene understanding within the 2D image domain. This advancement is made possible through the utilization of differential neural rendering techniques, such as NeRF (Neural Radiance Fields) \cite{mildenhall_nerf_2020}. Numerous NeRF-based approaches \cite{barron_mip-nerf_2021, barron_mip-nerf_2022, muller_instant_2022, barron_zip-nerf_2023, wang_f2-nerf_2023, goli_bayes_2023, tancik_nerfstudio_2023} have made significant advancements in terms of both quality and efficiency. Furthermore, some other techniques have empowered NeRF with improved scene understanding capabilities. Semantic NeRF \cite{zhi_-place_2021} first proposes the lifting of noisy 2D annotations to the 3D space based on NeRF. Significant progress has been achieved through the efforts of the following works \cite{vora_nesf_2021, yu_monosdf_2022, yang_learning_2021}. While these methods have shown promising results, they are currently limited to dense input viewpoints within indoor scenes and are only applicable to static environments. In this study, our focus lies in dynamic 3D scene understanding specifically tailored to urban settings, achieved by lifting 2D information to the 3D space.

\boldparagraph{Urban Scene Reconstruction}
Numerous studies have been conducted to reconstruct urban scenes using various methods. These methods can be categorized into three classes: point-based \cite{agarwal_building_2011, schonberger_structure--motion_2016}, mesh-based \cite{gallup_piecewise_2010, lafarge_hybrid_2013} and NeRF-based \cite{martin-brualla_nerf_2021, rematas_urban_2021, wimbauer_behind_2023, tancik_block-nerf_2022, zhang_nerflets_2023, guo_streetsurf_2023, lu_urban_2023}. While point-based and mesh-based methods demonstrate faithful reconstructions, they struggle to recover all aspects of the scene, especially when it comes to high-quality appearance modeling. In contrast, NeRF-based models allow for reconstructing scene appearance and enable high-quality rendering of novel viewpoints. However, these approaches are primarily designed for static scenes, lacking the ability to handle dynamic urban environments. In this study, our focus lies in addressing the challenges of dynamic urban scenes.

Several methods have also been developed to address the reconstruction of dynamic urban scenes. Many of these approaches rely on the availability of accurate 3D bounding boxes for moving objects in order to separate the dynamic elements from the static components, as seen in NSG \cite{ost_neural_2021}, MARS \cite{wu_mars_2023} and UniSim \cite{yang_unisim_nodate}. PNF \cite{kundu_panoptic_2022} takes a different approach by leveraging monocular-based 3D bounding box predictions and proposes a joint optimization of object poses during the reconstruction process. However, our experimental observations indicate that the straightforward optimization of object poses yields unsatisfactory results due to the absence of physical constraints. Another method, SUDS \cite{turki_suds_2023}, avoids the use of 3D bounding boxes by grouping the scene based on learned feature fields. However, the accuracy of this approach lags behind. In parallel, the concurrent work EmerNeRF \cite{yang_emernerf_2023} follows a similar idea to SUDS by decomposing the scene purely into static and dynamic components. In our research, we possess the capability to further decompose individual dynamic objects within the scene and estimate their motion.

\begin{figure*}
  \centering
  \includegraphics[width=0.98\linewidth]{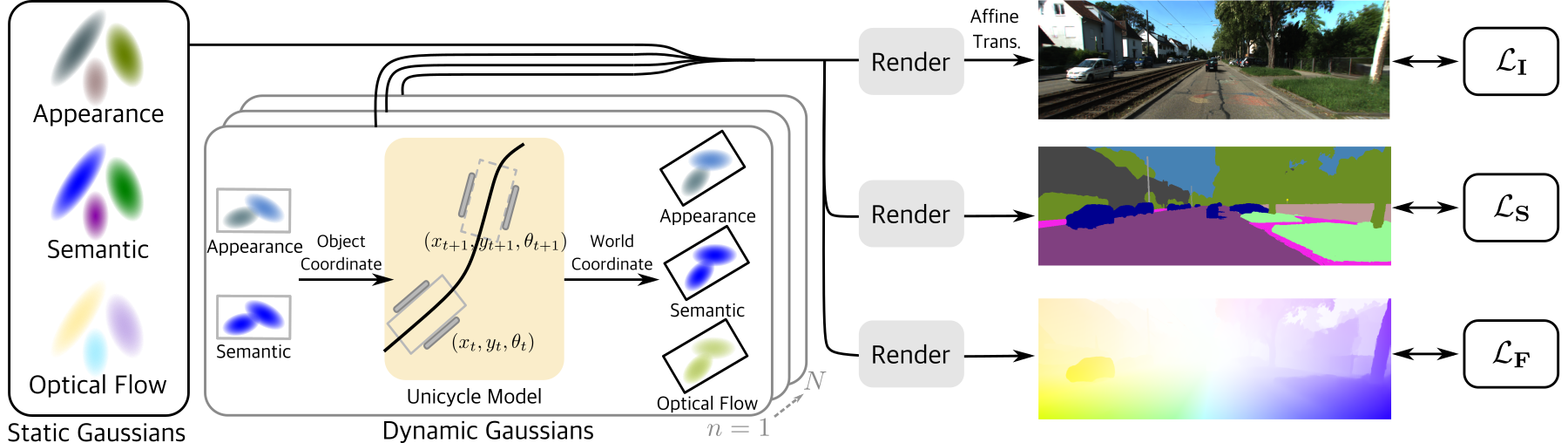}  
  \caption{\textbf{Method Overview.} We decompose the scene into static regions and $N$ rigidly moving dynamic objects. Each dynamic object is represented using 3D Gaussians in its canonical space and then transformed to the world coordinates based on transformations constrained by a unicycle model. We use $N$ unicycle models of different parameters to individually represent the motion of $N$ dynamic objects. Each 3D Gaussian encompasses information about appearance and semantics, whereas the optical flow can be obtained by calculating the Gaussian center's motion, enabling the rendering of RGB images, semantic maps, and optical flow within a unified model. Our method is supervised using RGB images, noisy 2D semantic labels, and noisy optical flow, denoted as $\cL_{\bI}$, $\cL_{\bS}$, and $\cL_{\bF}$, respectively. }
  \label{fig:method}
\vspace{-0.3cm}
\end{figure*}

\boldparagraph{Gaussian Splatting} 
3D Guassians are demonstrated as a powerful scene representation for novel view synthesis. While the original 3D Gaussian Splatting \cite{kerbl_3d_nodate} primarily focuses on static scenes, subsequent research has extended this approach to handle dynamic scenes. Dynamic 3D Gaussians \cite{luiten_dynamic_2023} necessitates a substantial number of training views accompanied by ground truth masks. Other studies \cite{xu_4k4d_2023, yang_deformable_2023, yang_real-time_2023, zielonka_drivable_2023} have also attempted to decompose 3D Gaussians into static and dynamic components, without further decomposing multiple dynamic objects. In our work, we strive to achieve the decomposition of each individual dynamic object while being capable of learning such decomposition from sparse urban images and noisy labels.

\section{Method}
\figref{fig:method} illustrates our proposed method, HUGS. Our algorithm takes as input posed images of a dynamic urban scene. We decompose the scene into static and dynamic 3D Gaussians, with the motion of dynamic vehicles being modeled via a unicycle model. The 3D Gaussians represent not only appearance but also semantic and flow information, allowing for rendering the RGB images, semantic labels, as well as optical flow through volume rendering. 

\subsection{Decomposed Scene Representation}
We assume that the scene is composed of static regions and a total of $N$ dynamic vehicles exhibiting rigid motions. Static regions are represented using static Gaussians in the world coordinate system. Each of the $N$ dynamic vehicles is modeled using dynamic Gaussians in a canonical coordinate system along with a set of rigid transformations $\{(\bR_t^n, \bt_t^n)\}_{t=1}^{T}$ with $t$ denoting the timestamp.

\boldparagraph{Static and Dynamic 3D Gaussians} Following Gaussian Splatting \cite{kerbl_3d_nodate}, we model both static and dynamic regions using 3D Gaussians. Each Gaussian is defined by a 3D covariance matrix $\bSigma\in \nR^{3 \times 3}$ and a 3D position $\mu \in \nR^{3}$, as well as an opacity $\alpha\in\nR^{+}$:
\begin{equation}
G(\bx ) =  \alpha \exp \left(-\frac{1}{2} (\bx-\mu)^T \bSigma^{-1} (\bx-\mu) \right)
\end{equation}
In addition, each Gaussian represents a color vector $\bc \in \nR^{3}$ parameterized as SH coefficients. In this work, we propose to additionally model semantic logits $\bs \in \nR^{S}$ of each 3D Gaussian, allowing for rendering 2D semantic labels. Furthermore, we can naturally obtain a rendered optical flow $\bff_{t_1 \rightarrow t_2 }\in \nR^2$ for each 3D Gaussian by projecting the 3D position $\mu$ to the image space at two different timestamps, $t_1$ and $t_2$, and calculating the motion.

\boldparagraph{Unicycle Model} We parameterize the transformations $(\bR_t, \bt_t)$ following the unicycle model\footnote{While it is more accurate to model vehicles using a bicycle model, we observe that using the simpler unicycle model is sufficient for our task.}. The state of a unicycle model is parameterized by three elements: $(x_t, y_t, \theta_t)$, where $x_t$ and $y_t$ represent the first two axes of $\bt$ with $\bt_t = [x_t,y_t,z_t]$, and $\theta_t$ is the yaw angle of $\bR_t$.
To adapt the continuous unicycle model to discrete frames, we derive the calculus of the unicycle model for the vehicle transition from timestamp $t$ to $t+1$ as follows:
\begin{align}
    x_{t+1} & = x_t + \frac{v_t}{\omega_t} (\sin \theta_{t+1} - \sin \theta_t)  \nonumber \\ 
    y_{t+1} & = y_t - \frac{v_t}{\omega_t} (\cos \theta_{t+1} - \cos \theta_t)  \label{eq:unicycle}  \\ 
    \theta_{t+1} & = \theta_t + \omega_t \nonumber
\end{align}
Here, $v_t$ represents the forward velocity, and $\omega_t$ is the angular velocity. This model integrates physical constraints when compared to directly optimizing the transformations of dynamic vehicles at every frame independently, thus enabling smoother motion modeling of moving objects and making them less prone to local minima.

While it is possible to define an initial state $(x_1, y_1, \theta_1)$ and derive the following states recursively based on velocities, $v_t$ and $\omega_t$, such a recursive parameterization is challenging to optimize.
In practice, we define a set of trainable states $\{(x_t, y_t, \theta_t)\}_{t=1}^{T}$ along with trainable velocities $\{v_t, \omega_t\}_{t=1}^{T-1}$,
and add a regularization term to ensure that the vehicle's states adhere to the characteristics of a unicycle model in \eqref{eq:unicycle}. The regularization terms will be described in \secref{sec:loss}. 
Additionally, we model the vertical locations of the vehicle, $\{z_t\}_{t=1}^{T}$, as optimizable parameters.

\subsection{Holistic Urban Gaussian Splatting}

Given the HUGS representation specified above, we are able to render images, semantic maps and optical flow to supervise the model or make predictions at inference time. We now elaborate on the rendering of each modality.

\boldparagraph{Novel View Synthesis}
The combination of static and dynamic Gaussians can be sorted and projected onto the image plane via $\alpha$-blending:
\begin{equation}
\pi: \quad \bC = \sum_{i \in \mathcal{N}} \bc_i \alpha'_i \prod_{j=1}^{i-1}(1-\alpha'_j)
\label{eq:alpha_blender}
\end{equation}
Here, $\alpha'_j$ is determined by the projected 2D Gaussian and the 3D opacity $\alpha$, see supplement for details.

In contrast to single-object scenes, urban scenes typically involve more complex lighting conditions and the images are usually captured with auto white balance and auto exposure. NeRF-based methods \cite{martin-brualla_nerf_2021} typically feed a per-frame appearance embedding along with the 3D positions into a neural network to compute the color, thereby compensating exposure. However, when working with 3D Gaussians, there is no neural network capable of processing appearance embeddings.  Inspired by Urban Radiance Field \cite{rematas_urban_2021}, we generate an exposure affine matrix for each camera by mapping the camera's extrinsic parameters to an affine matrix $\bA \in \nR^{3 \times3}$ and vector $\bb \in \nR^3$ via a small MLP:
\begin{equation}
\Tilde{\bC} = \bA \times \bC + \bb
\end{equation}
We demonstrate that modeling the exposure improves rendering quality in the experimental section.

\begin{figure}
     \centering
     \small 
     \setlength{\tabcolsep}{0pt}
     \def\mywidth{4cm}
     \begin{tabular}{P{\mywidth}P{\mywidth}}

     \includegraphics[width=\mywidth]{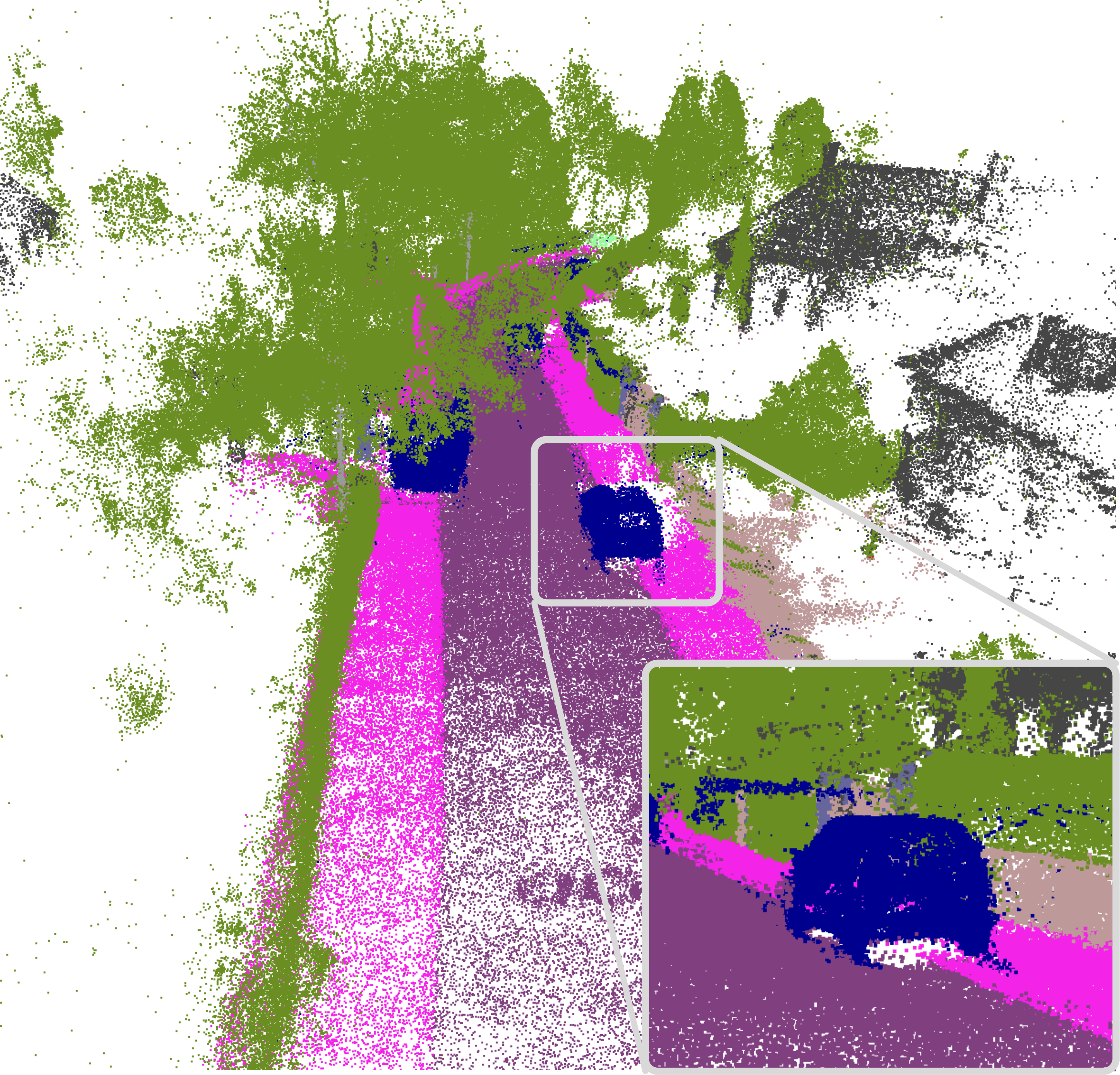}  &
     \includegraphics[width=\mywidth]{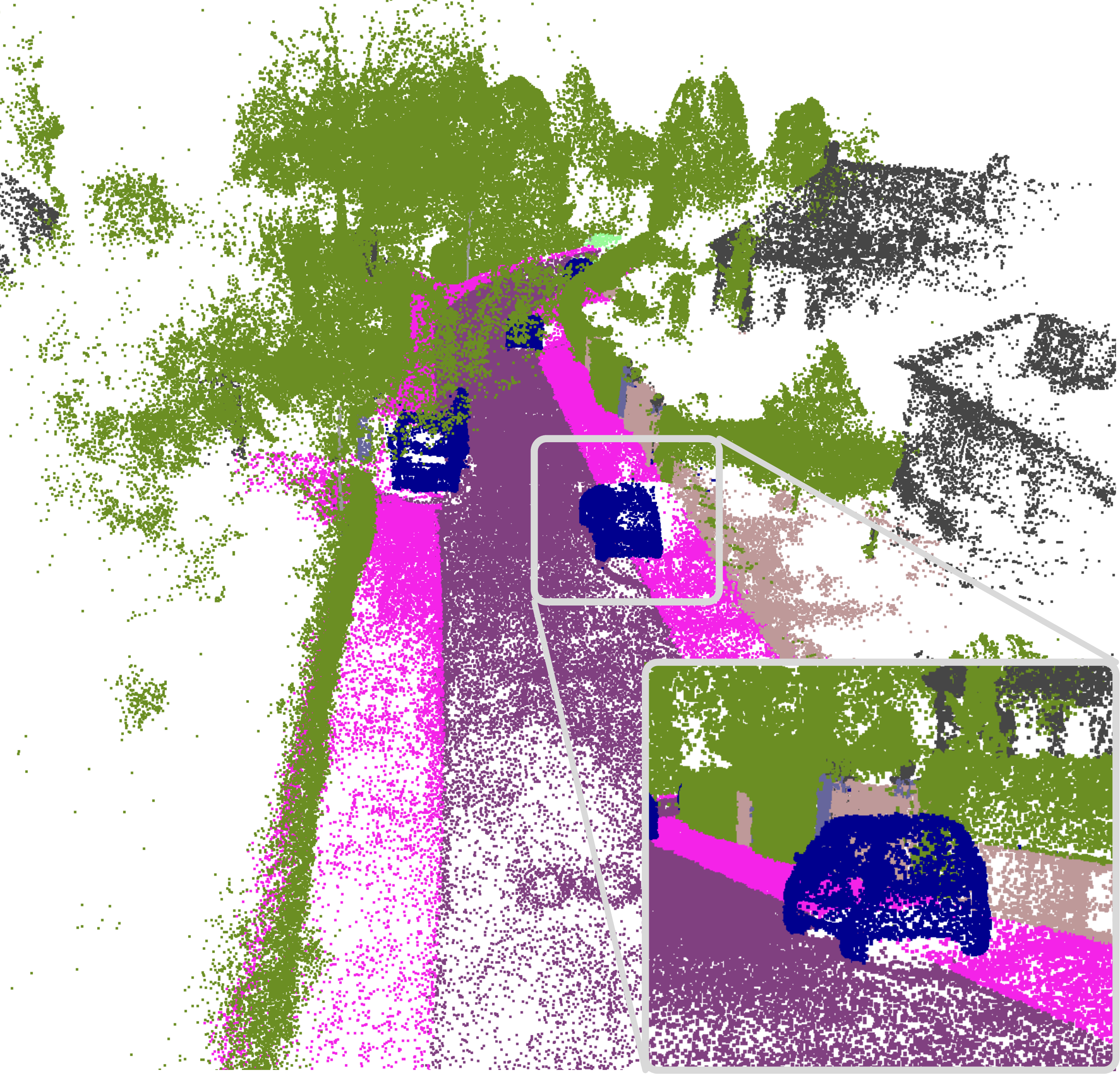} \\

     \end{tabular}
     \vspace{-0.2cm}
     \caption{\textbf{3D Semantic Reconstruction.} Comparison between applying softmax to accumulated 2D semantic logits (left) and to 3D semantic logits (right). Normalizing semantic logits in 3D space clearly reduces floaters and yields better 3D semantic reconstruction than the 2D normalization counterpart.}
     \vspace{-0.3cm}
\label{fig:softmax_cmp}
\end{figure}

\boldparagraph{Semantic Reconstruction}
Similarly to \eqref{eq:alpha_blender}, we can obtain 2D semantic labels via $\alpha$-blending based on the 3D semantic logit $\bs$:
\begin{equation}
\pi: \quad \bS = \sum_{i \in \mathcal{N}} \text{softmax}(\bs_i) \alpha'_i \prod_{j=1}^{i-1}(1-\alpha'_j)
\label{eq:semantic_render}
\end{equation}
Note that we perform the softmax operation on 3D semantic logits $\bs_i$ prior to $\alpha$ blending, in contrast to most existing methods that apply softmax to 2D semantic logits $\bar{\bS}$ obtained by accumulating unnormalized 3D semantic logits $\bs_i$~\cite{zhi_-place_2021, fu_panoptic_2022}. As shown in \figref{fig:softmax_cmp}, applying softmax in 2D space leads to noisy 3D semantic labels. This is due to the fact that 2D space softmax can produce accurate 2D semantics by adjusting the scale of the 3D semantic logits,  allowing a single sampled point with a substantial logit value to significantly influence the volume rendering outcome.
For example, an undesired floating point labeled with ``car'' may not be penalized despite the target rendered label is ``tree'', as long as there is a 3D Gaussian providing a large logit value of ``tree'' along this ray. Our solution instead removes such floaters by normalizing logits in 3D space. See supplement for more quantitative and qualitative details. %

\boldparagraph{Optical Flow} The 3D Gaussian representation also enables the rendering of optical flow. 
Given two timestamps $t_1$ and $t_2$, we first calculate the optical flow of each 3D Gaussian's center $\mu$ as $\bff_{t_1\rightarrow t_2}$. Specifically, we project $\mu$ to the 2D image space based on the camera's intrinsic and extrinsic parameters: 
\begin{equation}
    \mu'_1=\bK{[\bR_{t_1}^{\text{cam}};\bt_{t_1}^{\text{cam}}]}\mu, \quad 
    \mu'_2=\bK{[\bR_{t_2}^{\text{cam}};\bt_{t_2}^{\text{cam}}]}\mu,
\end{equation}
and then calculate the motion vector as $\bff_{t_1\rightarrow t_2}=\mu'_2-\mu'_1$.
Next, we render the optical flow via accumulate the optical flows via volume rendering:
\begin{equation}
\pi: \quad \bF = \sum_{i \in \mathcal{N}} \bff_i \alpha'_i \prod_{j=1}^{i-1}(1-\alpha'_j)
\label{eq:flow_render}
\end{equation}
Note that this rendering process assumes that any pixel of a 2D Gaussian splat shares the same optical flow direction as the corresponding Gaussian center but with scaled magnitude. While this is indeed a simplified approximation, we observe this to work well in practice.

In our experiments, we demonstrate that supervising the rendered optical flow with pseudo ground truth helps to improve the performance of the geometry in terms of rendered depth maps. This is due to the fact that flow provides explicit pixel correspondences, which is inherently supervising the underlying surface location.

\subsection{Loss Functions}
\label{sec:loss}
We leverage pre-trained recognition models to provide noisy 2D semantic and instance predictions, noisy 2D optical flow, as well as noisy 3D tracking results. These easy-to-obtain predictions are critical to enable RGB-only holistic scene understanding in both 2D and 3D space, without relying on LiDAR input or 3D semantic supervision.

\boldparagraph{Image-based Losses}
Our model is supervised with the ground truth images using a combination of L1 and SSIM losses. Let $\Tilde{\bI}$ denote the rendered image and $\hat{\bI}$ the ground truth, our rendering loss is defined as follows:
\begin{equation}
    \cL_{\bI} = (1-\lambda_{SSIM}) \| \hat{\bI} - \Tilde{\bI} \|_1 + \lambda_{SSIM}\text{SSIM}(\hat{\bI}, \Tilde{\bI})
\end{equation}
We additionally apply the cross-entropy loss to the rendered semantic label \wrt pseudo-2D semantic segmentation ground truth $\hat{\bS}$:
\begin{equation}
    \cL_{\bS} = - \sum_{k=0}^{S-1} \hat{\bS}_k \log (\bS_k) 
\end{equation}
Similarly, we leverage pseudo optical flow ground truth $\hat{\bF}$ to supervise the rendered optical flow using:
\begin{equation}
    \cL_{\bF} = \| \hat{\bF} - \bF \|_1 
\end{equation}
While 3D Gaussians can enable the rendering of optical flow without any supervision, we observe artifacts in the rendered flow without supervision. Further, the optical flow supervision yields an improvement in the depth maps as shown in our ablation study.

\boldparagraph{Unicycle Model Losses}
We use a unicycle model to guide the noisy 3D bounding box predictions:
\begin{equation}
\label{eq:L_3dbox}
\cL_{\bt} = \sum_t \Vert x_t - \hat{x}_t \Vert_2 + \sum_t \Vert y_t - \hat{y}_t\Vert_2
\end{equation}
where $\hat{x}_t$ and $\hat{y}_t$ are the $x$ and $y$ locations of a noisy 3D bounding box at timestamp $t$.

As mentioned earlier, we parameterize the vehicle's states $(x_t, y_t, \theta_t)$ and the velocities $v_t, \omega_t$ as learnable parameters. Hence, we add the following regularization to make the states adhere to the unicycle model as follows:
\begin{align}
\cL_{uni} = & \sum_t \| x_{t+1} - x_t - \frac{v_t}{\omega_t} (\sin \theta_{t+1} - \sin \theta_t) \| + \nonumber \\
           &   \sum_t\|  y_{t+1} - y_t + \frac{v_t}{\omega_t} (\cos \theta_{t+1} - \cos \theta_t) \| +    \nonumber\\
           &  \sum_t \|  \theta_{t+1} - \theta_t  - \omega_t \| 
            \label{eq:uni_restrain}
\end{align}
In addition, we regularize the acceleration of the forward velocity $v_t$ and angular velocity $\omega_t$ to be smooth:
\begin{align}
\cL_{reg} = & \sum_t \Vert v_{t+1} + v_{t-1} - 2v_t \Vert_2 + \nonumber \\
            & \sum_t \Vert \theta_{t+1} + \theta_{t-1} - 2\theta_t \Vert_2
            \label{eq:uni_reg}
\end{align}

The total loss can be summarized as follows:
\begin{equation}
\cL = \cL_\bI + \lambda_\bS \cL_\bS +  \lambda_\bF \cL_\bF + \lambda_\bt \cL_\bt + \lambda_{uni} \cL_{uni} + \lambda_{reg} \cL_{reg}
\end{equation}

\subsection{Implementation Details}
\boldparagraph{Initialization}
While 3D Gaussian Splatting is not highly sensitive to the initialization, better initialization can yield better performance.
We utilize the dense point cloud obtained from COLMAP for initialization by default. When the ego-vehicle is static, we use random initialization.

\boldparagraph{Pseudo-GTs}
We utilize InverseForm \cite{borse_inverseform_2021} to generate pseudo ground truth for semantic segmentation.
For initializing the unicycle model, we employ a monocular-based method, QD-3DT \cite{hu_monocular_2021}, to acquire pseudo ground truth for 3D bounding boxes and tracking IDs at each training view. For optical flow, we use Unimatch \cite{xu_unifying_2023} to obtain pseudo ground truth. 

\boldparagraph{Training}
We train the model for 30,000 iterations on dynamic scenes. For the KITTI-360 leaderboard, we perform early stopping at 15,000 iterations. 
Following \cite{kerbl_3d_nodate}, we adopt the approach of setting the weight parameter $\lambda_{SSIM}$ to 0.2. Furthermore, we assign weights $\lambda_\bS$ and $\lambda_\bF$ as 0.01, while $\lambda_\bt$, $\lambda_{uni}$ and $\lambda_{reg}$ are set as 0.1. The learning rate of the unicycle model parameters progressively decreases during training. 

\boldparagraph{Time Consuming}
Our approach can converge within 30 minutes and achieve inference at a speed of approximately 93 fps on a single NVIDIA RTX 4090. While NSG and MARS inference at a speed of less than 1 fps. A speed breakdown of our method is provided in the supplement.

\section{Experiments}

\begin{figure*}[t!]
     \centering
     \small 
     \setlength{\tabcolsep}{0pt}
     \def\mywidth{4.2cm}
     \begin{tabular}{P{0.5cm}P{\mywidth}P{\mywidth}P{\mywidth}P{\mywidth}}
     \multirow{2}{*}{\rotatebox{90}{KITTI~~~~}} &
     \includegraphics[width=\mywidth]{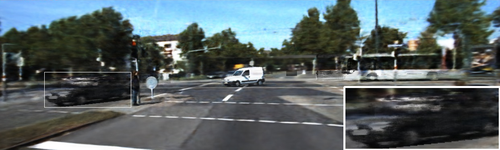}  & 
     \includegraphics[width=\mywidth]{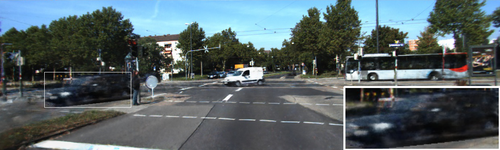} &
     \includegraphics[width=\mywidth]{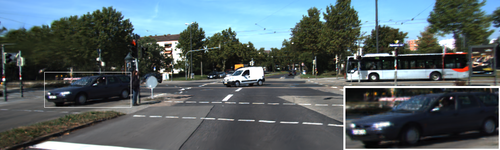} &
     \includegraphics[width=\mywidth]{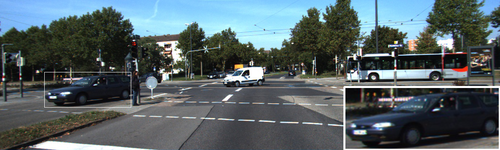} \\
     
     &
     \includegraphics[width=\mywidth]{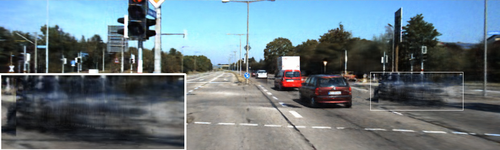}  & 
     \includegraphics[width=\mywidth]{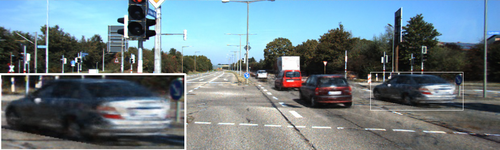} &
     \includegraphics[width=\mywidth]{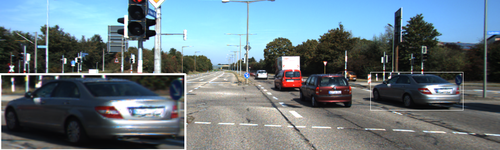} & 
     \includegraphics[width=\mywidth]{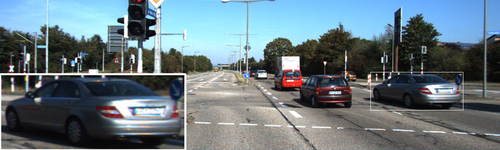} \\

     \multirow{2}{*}{\rotatebox{90}{vKITTI~~~~}} &
     \includegraphics[width=\mywidth]{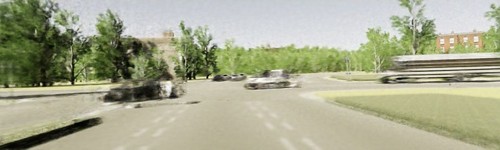}  & 
     \includegraphics[width=\mywidth]{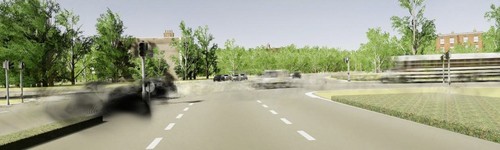} &
     \includegraphics[width=\mywidth]{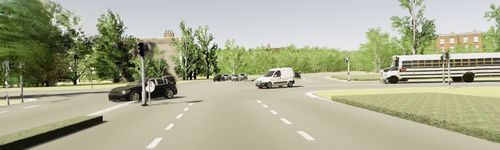} & 
     \includegraphics[width=\mywidth]{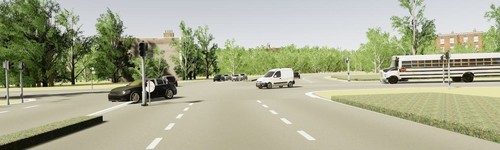} \\
     
     &
     \includegraphics[width=\mywidth]{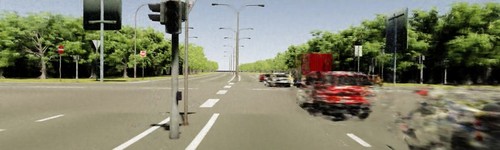}  & 
     \includegraphics[width=\mywidth]{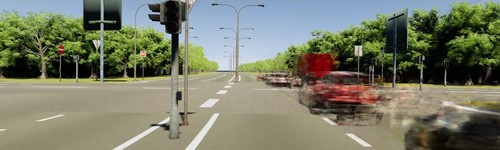} &
     \includegraphics[width=\mywidth]{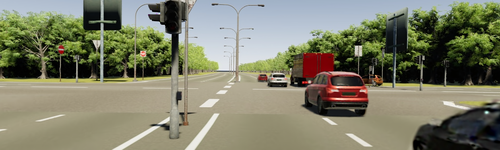} & 
     \includegraphics[width=\mywidth]{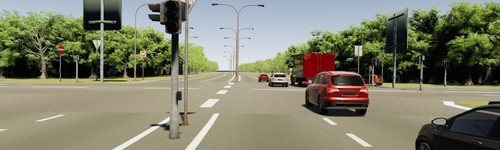} \\
     \rotatebox{0}{} & \rotatebox{0}{NSG} & \rotatebox{0}{MARS} & \rotatebox{0}{Ours}  & \rotatebox{0}{GT}  \\

     \end{tabular}\vspace{-0.2cm}
     \caption{\textbf{Qualitative Comparison} on KITTI and vKITTI. We use monocular-based 3D bounding box predictions for KITTI, and manually jittered 3D bounding boxes for vKITTI. We zoom in on a patch of a dynamic object for each KITTI scene.}
    \vspace{-0.2cm}
\label{fig:vkitti_noise}
\end{figure*}

\begin{table*}[]
\centering
\small
\setlength{\tabcolsep}{4pt}
\begin{tabular}{@{\extracolsep{2pt}}lcccccccccccc@{}} 
\toprule
\multicolumn{1}{c}{} & \multicolumn{3}{c}{KITTI Scene02} & \multicolumn{3}{c}{KITTI Scene06} & \multicolumn{3}{c}{vKITTI Scene02} & \multicolumn{3}{c}{vKITTI Scene06} \\
& PSNR$\uparrow$  & SSIM$\uparrow$  & LPIPS$\downarrow$  & PSNR$\uparrow$  & SSIM$\uparrow$  & LPIPS$\downarrow$ & PSNR$\uparrow$  & SSIM$\uparrow$  & LPIPS$\downarrow$  & PSNR$\uparrow$  & SSIM$\uparrow$  & LPIPS$\downarrow$  \\
\cline{2-4} \cline{5-7} \cline{8-10}  \cline{11-13} 
NSG \cite{ost_neural_2021} 
& 23.00 & 0.664 & 0.373 & 23.78 & 0.717 & 0.234 & 21.40 & 0.689 & 0.376 & 20.60 & 0.719 & 0.255 \\
MARS \cite{wu_mars_2023} 
& 23.30 & 0.731 & 0.139 & 25.09 & 0.856 & 0.083 & 22.67 & 0.882 & 0.128 & 21.67 & 0.856 & 0.134 \\
Ours 
& \textbf{25.42} & \textbf{0.821} & \textbf{0.092} & \textbf{28.20} & \textbf{0.919} & \textbf{0.027} & \textbf{26.21} & \textbf{0.911} & \textbf{0.040} & \textbf{26.65} & \textbf{0.921} & \textbf{0.030} \\
\bottomrule
\end{tabular}
\vspace{-0.2cm}
\caption{\textbf{Novel View Synthesis on Dynamic Scenes} with predicted or noisy 3D trackings.}
\label{tab:dynamic_noisy}
\vspace{-0.2cm}
\end{table*}

\boldparagraph{Datasets}
We perform a range of experiments to assess the performance of our model across various tasks, such as novel view synthesis, novel semantic synthesis, and 3D semantic reconstruction. These experiments are conducted using the KITTI \cite{geiger_are_2012}, Virtual KITTI 2 (vKITTI) \cite{cabon_virtual_2020}, and KITTI-360 datasets \cite{liao_kitti-360_2022}. We apply 50\% dropout rate following existing evaluation protocols \cite{liao_kitti-360_2022, wu_mars_2023} on all of these datasets.

\boldparagraph{Baselines}
We evaluate the dynamic scene novel view synthesis task by comparing our method with NSG \cite{ost_neural_2021} and MARS \cite{wu_mars_2023}, which are two open-source methods for dynamic urban scenes. Additionally, we compare the static novel view appearance and semantic synthesis task with mip-NeRF \cite{barron_mip-nerf_2021}, PNF \cite{kundu_panoptic_2022}, and MARS \cite{wu_mars_2023}. Furthermore, we assess the quality of 3D semantic scene reconstruction by comparing it with Semantic Nerfacto \cite{tancik_nerfstudio_2023}.

\boldparagraph{Evaluation Metrics} For \textit{novel view synthesis}, we adopt the default setting for quantitative assessments, including the evaluation of PSNR, SSIM and LPIPS \cite{zhang_unreasonable_2018}.
Regarding \textit{novel view semantic synthesis}, we follow KITTI-360 \cite{liao_kitti-360_2022}, which reports the mean Intersection over Union on class (mIoU$_\text{cls}$) and category (mIoU$_\text{cat}$), respectively.  Further, we evaluate our performance on \textit{3D Semantic Segmentation} against a ground truth semantic LiDAR point cloud, measuring both geometric reconstruction quality and semantic accuracy. The geometric quality is evaluated as the chamfer distance between two point clouds, including completeness and accuracy, whereas the semantic accuracy is also measured using mIoU$_\text{cls}$.
In our ablation study, we evaluate \textit{3D tracking} performance by measuring the rotation and translation error $e_\bR$ and $e_\bt$ of our optimized 3D bounding boxes wrt. the ground truth.

\subsection{Novel View Synthesis}

We first evaluate HUGS for novel view synthesis on various datasets including dynamic and static scenes. For dynamic scenes, we leverage noisy 3D bounding box predictions as input, instead of using the ground truth. Despite not being our main focus, we include a comparison of using ground truth 3D bounding boxes in the supplement.

\boldparagraph{Dynamic Scene with Noisy 3D Bounding Boxes}
Following \cite{ost_neural_2021, wu_mars_2023}, we evaluate our performance on dynamic scenes of the KITTI and vKITTI datasets. In contrast to these methods that leverage ground truth poses, we investigate a more practical scenario where the bounding boxes are generated by a monocular-based 3D tracking algorithm, QD-3DT~\cite{hu_monocular_2021}, in \tabref{tab:dynamic_noisy}. Here, the predicted 3D bounding boxes are only provided for training views, as testing views should not be used as inputs for the tracking model. In experiments where the unicycle model is not utilized, the bounding boxes of testing views are obtained through linear interpolation from neighbour training views. Where the unicycle model is used, the bounding boxes of testing views are computed using \eqref{eq:unicycle}. For vKITTI, there is no pre-trained monocular tracking algorithm. We hence jitter the ground truth poses to simulate noisy monocular predictions, with an average noise of 0.5 meters in translation and 5 degrees in rotation. Our model's robustness wrt. various levels of noise will be analyzed in the ablation study. 

\tabref{tab:dynamic_noisy} demonstrate that our method consistently outperforms against the baselines. Note, that QD-3DT yields reasonable predictions on the KITTI dataset\footnote{In fact, following the evaluation protocol of MARS, the sequences we evaluate on are used as training sequences for QD-3DT.}. Hence, NSG and MARS reconstruct the dynamic objects reasonably well, but with more blurriness and artifacts (see \figref{fig:vkitti_noise}), as they do not model the optimization of the object poses. In contrast, our method allows for reconstructing dynamic objects with sharp details, not only in cases of minor pose error on the KITTI dataset but also on the vKITTI dataset with more severe noise.

\begin{figure}[t!]
     \centering
     \small 
     \setlength{\tabcolsep}{0pt}
     \def\mywidth{2.0cm}
     \begin{tabular}{P{\mywidth}P{\mywidth}P{\mywidth}P{\mywidth}}
     \includegraphics[width=\mywidth]{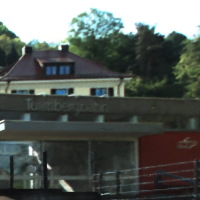}  
     & \includegraphics[width=\mywidth]{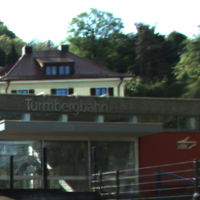}
     & \includegraphics[width=\mywidth]{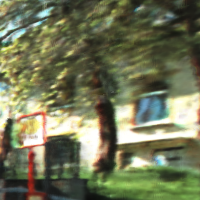}
     & \includegraphics[width=\mywidth]{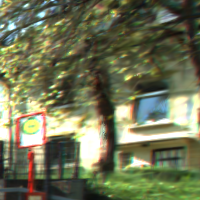} \\
     \includegraphics[width=\mywidth]{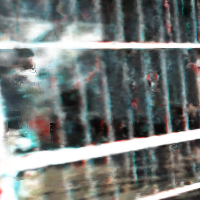}  
     & \includegraphics[width=\mywidth]{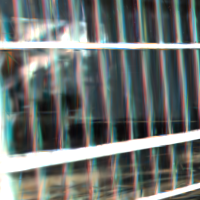}
     & \includegraphics[width=\mywidth]{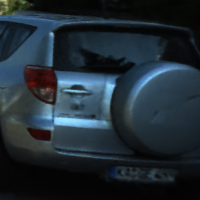}
     & \includegraphics[width=\mywidth]{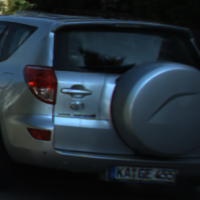} \\
     \rotatebox{0}{MARS} & \rotatebox{0}{Ours} & \rotatebox{0}{MARS} & \rotatebox{0}{Ours}\\
     \end{tabular}
     \vspace{-0.3cm}
     \caption{\textbf{Details Qualitative Comparison} with MARS on KITTI-360 Leaderboard.}
     \vspace{-0.2cm}
\label{fig:compare_mars_detail}
\end{figure}

\begin{table}[t]
\centering
\small
\setlength{\tabcolsep}{0.6pt}
\def\mywidth{1.3cm}
\begin{tabular}{lP{1.2cm}P{1.2cm}P{1.2cm}|P{\mywidth}P{\mywidth}}
\toprule
& PSNR$\uparrow$     & SSIM$\uparrow$  & LPIPS$\downarrow$   & mIoU$_\text{cls}\uparrow$  & mIoU$_\text{cat}\uparrow$       \\
\cline{2-6}
mip-NeRF \cite{barron_mip-nerf_2021}
& 21.54                 & 0.778          & 0.365          & 48.25          & 67.47             \\
PNF \cite{kundu_panoptic_2022}   
& 22.07                 & 0.820          & 0.221          & \textbf{73.06} & 84.97       \\
MARS \cite{wu_mars_2023}
& 23.09                 & 0.857          & 0.174          & -              & - \\
Ours     
& \textbf{23.38}        & \textbf{0.870} & \textbf{0.121} & 72.65          & \textbf{85.64}        \\ 
\bottomrule
\end{tabular}
\vspace{-0.3cm}
\caption{\textbf{Novel View Semantic and Appearance Synthesis} on KITTI-360.}
\label{tab:leaderboard_kitti360}
\vspace{-0.3cm}
\end{table}

\boldparagraph{Static Scene Leaderboard} We further evaluate our performance on the KITTI-360 leaderboard, which contains 5 static sequences. Our method achieves state-of-the-art performance on the leaderboard as in \tabref{tab:leaderboard_kitti360} (left), demonstrating the effectiveness of the 3D Gaussian representation in modeling complex urban scenes. As we will discuss in the ablation study, incorporating the affine transform to model camera exposure is important for reaching high fidelity. \figref{fig:compare_mars_detail} shows the qualitative comparison of our proposed method to another top-ranking method, MARS, on the leaderboard. 

\subsection{Semantic and Geometric Scene Understanding}

Next, we evaluate our model on various semantic and geometric scene understanding tasks on the KITTI-360 dataset.

\begin{figure}[t!]
     \centering
     \small 
     \setlength{\tabcolsep}{0pt}
     \def\mywidth{4.2cm}
     \begin{tabular}{P{\mywidth}P{\mywidth}}
     \includegraphics[width=\mywidth]{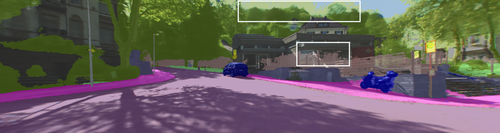}  & \includegraphics[width=\mywidth]{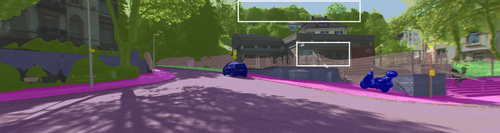} \\
     \includegraphics[width=\mywidth]{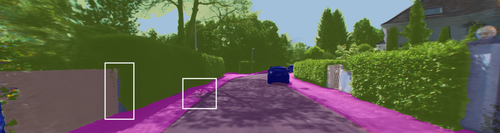}  & \includegraphics[width=\mywidth]{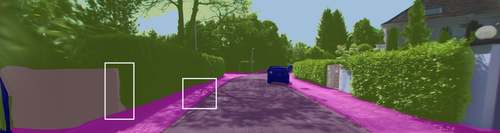} \\
     \includegraphics[width=\mywidth]{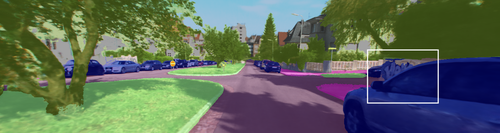}  & \includegraphics[width=\mywidth]{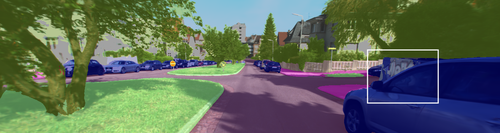} \\
    \rotatebox{0}{PNF} & \rotatebox{0}{Ours}
     \end{tabular}
    \vspace{-0.3cm}
     \caption{\textbf{Qualitative Comparison} with PNF on KITTI-360 Leaderboard.}
     \vspace{-0.2cm}
\label{fig:compare_pnf}
\end{figure}

\boldparagraph{Novel View Semantic Synthesis}
Our holistic representation also enables novel view semantic synthesis. Hence, we submit our novel view semantic synthesis performance to the KITTI-360 leaderboard for comparison as well, see \tabref{tab:leaderboard_kitti360} (right). Despite not leveraging category-level prior as done in previous work~\cite{kundu_panoptic_2022}, our approach achieves comparable performance to the SOTA~\cite{kundu_panoptic_2022} as shown in \figref{fig:compare_pnf}.

\boldparagraph{3D Semantic Scene Reconstruction}
While existing 2D-to-3D semantic lifting methods solely evaluate their performance in the 2D image space, we further evaluate our performance in the 3D space to examine the underlying 3D geometry. To this goal, we leverage the ground truth LiDAR points provided by the KITTI-360 dataset for evaluation.
With each Gaussian possessing semantic information, we can obtain a semantic point cloud by extracting the Gaussian's center $\mu$ and its semantic label. 
We evaluate the geometric quality and semantic accuracy of this semantic point cloud in \tabref{tab:semantic_3d}. We compare our method with Semantic Nerfacto \cite{tancik_nerfstudio_2023}, a Semantic NeRF implemented using a more advanced backbone, as the state-of-the-art novel view semantic synthesis method, PNF, in \tabref{tab:leaderboard_kitti360} is not open-source. For this baseline, we extract a semantic point cloud by specifying a threshold to the density field.
While Semantic Nerfacto enables rendering faithful 2D semantic labels as shown in the supplement, the underlying 3D semantic point cloud is significantly worse in comparison. The Gaussian based representation instead allows for extracting a much more accurate semantic point cloud in comparison. 
\begin{table}[t]
\centering
\small 
\begin{tabular}{lcccc}
\toprule 
& acc.$\downarrow$ & comp.$\downarrow$  & mIoU$_\text{cls}\uparrow$ \\
\cline{2-4}
Semantic Nerfacto
& 1.508 & 24.28 & 0.055 \\
Ours
& \textbf{0.233} & \textbf{0.214} & \textbf{0.505} \\ 
\bottomrule
\end{tabular}
\caption{\textbf{3D Semantic Reconstruction} on KITTI-360. Note that all metrics are calculated in 3D space. }
\label{tab:semantic_3d}
\vspace{-0.2cm}
\end{table}

\subsection{Scene Editing}

\begin{figure}[t!]
    \centering
    \small 
    \setlength{\tabcolsep}{0pt}
    \def\mywidth{4.2cm}
    \begin{tabular}{P{\mywidth}P{\mywidth}}
    \includegraphics[width=\mywidth]{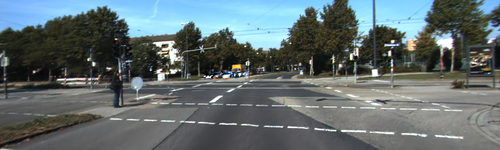}  &
    \includegraphics[width=\mywidth]{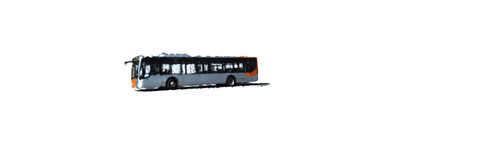} \\
    \includegraphics[width=\mywidth]{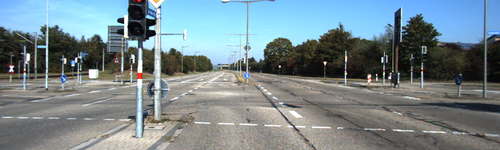}  &
    \includegraphics[width=\mywidth]{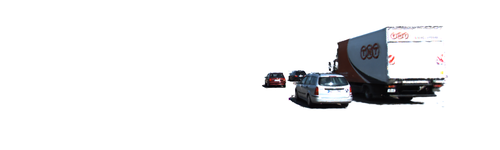} \\
    \rotatebox{0}{Background} & \rotatebox{0}{Foreground} \\
    \end{tabular}
    \vspace{-0.3cm}
    \caption{\textbf{Scene Decomposition} on KITTI. Our approach enables clear decomposition of foreground and background.}
    \vspace{-0.2cm}
    \label{fig:decompose}
\end{figure}

\begin{figure}[t!]
    \centering
    \small 
    \setlength{\tabcolsep}{0pt}
    \def\mywidth{4.2cm}
    \begin{tabular}{P{\mywidth}P{\mywidth}}
    \includegraphics[width=\mywidth]{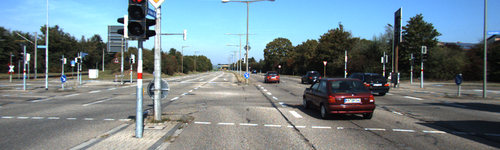}  &
    \includegraphics[width=\mywidth]{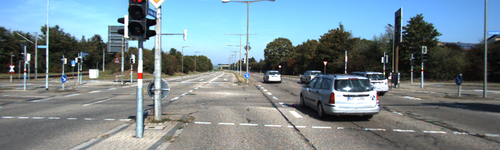} \\
    \includegraphics[width=\mywidth]{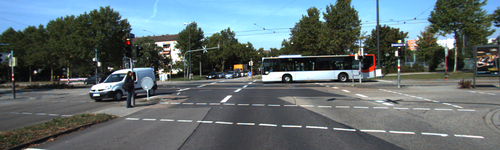}  &
    \includegraphics[width=\mywidth]{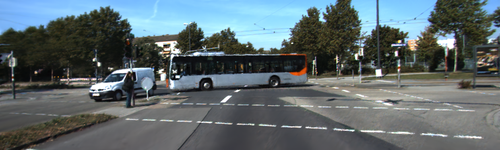} \\
    \includegraphics[width=\mywidth]{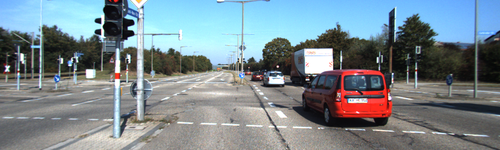}  &
    \includegraphics[width=\mywidth]{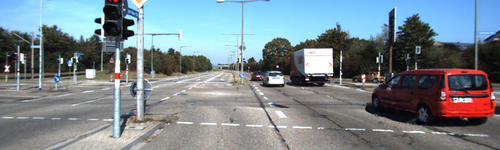} \\
    \rotatebox{0}{Original} & \rotatebox{0}{Edited} \\
    \end{tabular}
    \vspace{-0.3cm}
    \caption{\textbf{Scene Editing} on KITTI. Our decomposed scene representation enables replacing dynamic objects (1st row) and moving dynamic objects around (2nd \& 3rd rows).}
    \vspace{-0.2cm}
    \label{fig:editing}
\end{figure}

Our decomposed scene representation enables various downstream applications. Our method allows for decomposing foreground moving objects from the background as shown in \figref{fig:decompose}. Further, we can edit the scene by swapping dynamic objects, or manipulating their rotation and translations, see \figref{fig:editing}.

\begin{table*}[]
\centering
\small
\setlength{\tabcolsep}{1.2pt}
\begin{tabular}{@{\extracolsep{2pt}}lccccccccccccccc@{}} 
\toprule
\multicolumn{1}{c}{}                & \multicolumn{5}{c}{KITTI (5\% noise)} & \multicolumn{5}{c}{KITTI (10\% noise)} & \multicolumn{5}{c}{KITTI (20\% noise)}                             \\
& PSNR$\uparrow$  & SSIM$\uparrow$  & LPIPS$\downarrow$    & $e_\bR\downarrow$ & $e_\bt\downarrow$   & PSNR$\uparrow$  & SSIM$\uparrow$  & LPIPS$\downarrow$    & $e_\bR\downarrow$ & $e_\bt\downarrow$    &PSNR$\uparrow$  & SSIM$\uparrow$  & LPIPS$\downarrow$ & $e_\bR\downarrow$ & $e_\bt\downarrow$ \\
\cline{2-6} \cline{7-11} \cline{12-16} 
w/o opt., w/o uni.                    
& 23.83 & 0.878 & 0.062 & 0.031 & 0.027 & 22.16 & 0.861 & 0.079 & 0.063 & 0.106 & 20.28 & 0.835 & 0.101 & 0.125 & 0.425 \\
w/ opt.,  w/o uni.
& 24.80 & 0.897 & 0.038 & 0.022 & 0.051 & 22.75 & 0.879 & 0.056 & 0.054 & 0.130 & 20.56 & 0.855 & 0.081 & 0.135 & 0.612 \\
w/ opt.,  w/ uni. (Ours) 
& \textbf{28.78} & \textbf{0.928} & \textbf{0.023} & \textbf{0.017} & \textbf{0.022} & \textbf{26.66} & \textbf{0.908} & \textbf{0.032} & \textbf{0.037} & \textbf{0.035} & \textbf{23.59} & \textbf{0.875} & \textbf{0.061} & \textbf{0.081} & \textbf{0.176} \\
\bottomrule
\end{tabular}
\vspace{-0.2cm}
\caption{\textbf{Ablation Study on Dynamic Scenes} of KITTI.}
\vspace{-0.3cm}
\label{tab:ablation_dynamic}
\end{table*}

\subsection{Ablation Study}
We conduct ablation studies on dynamic and static scenes, respectively.

\boldparagraph{Dynamic Scene} As KITTI provides accurate 3D bounding box ground truth, we ablate the effectiveness of our unicycle model on KITTI by manually adding noise to the 3D bounding boxes and evaluate both the novel view synthesis results and the tracking performance, see \tabref{tab:ablation_dynamic}. In this experiment, we compare our full model to two variants, i.e., using the noises without optimization (w/o opt., w/o uni.), and performing na\"ive per-frame optimization without using the unicycle model (w/ opt., w/o uni.). The results validate the effectiveness of the unicycle model, which obviously improves the rendering quality and 3D tracking accuracy. Qualitative results in \figref{fig:comp_noisy} further verify the effectiveness of our unicycle model in enabling accuracy object reconstruction given noisy 3D bounding boxes.

\boldparagraph{Static Scene} We further study the effect of different components on three static scenes of KITTI-360 in \tabref{tab:ablation_static}. This allows us to ablate design choices without mixing up the impact of dynamic objects. The results indicate the significance of exposure modeling, which is particularly important for scenes with strong exposure variance. The semantic and flow losses have little contribution in improving novel view synthesis. It is rational as imposing a constraint on the semantic or flow does not necessarily contribute to appearance. However, note that incorporating the flow supervision clearly improves the underlying geometry, since optical flow provides explicit correspondence. See supplement for qualitative comparison.

\begin{table}[]
\centering
\small
\setlength{\tabcolsep}{2.1pt}
\begin{tabular}{@{\extracolsep{2pt}}lcccc@{}} 
\toprule
& PSNR$\uparrow$  & SSIM$\uparrow$  & LPIPS$\downarrow$ & Depth $\downarrow$  \\
\cline{2-5}
w/o Affine transform
& 24.18 & 0.827 & 0.083 & --\\
w/o $\cL_\bS$ 
& 24.47 & 0.831 & 0.081 & 0.892 \\
w/o $\cL_\bF$
& 24.45 & 0.831 & \textbf{0.080} & 1.031  \\
Ours 
& \textbf{24.52} & \textbf{0.833} & 0.081 & \textbf{0.872} \\
\bottomrule
\end{tabular}
\vspace{-0.2cm}
\caption{\textbf{Ablation Study on Static Scenes} on KITTI-360.}
\vspace{-0.2cm}
\label{tab:ablation_static}
\end{table}

\begin{figure}[t!]
     \centering
     \small 
     \setlength{\tabcolsep}{0pt}
     \def\mywidth{2.85cm}
     \begin{tabular}{P{\mywidth}P{\mywidth}P{\mywidth}}
     \includegraphics[width=\mywidth]{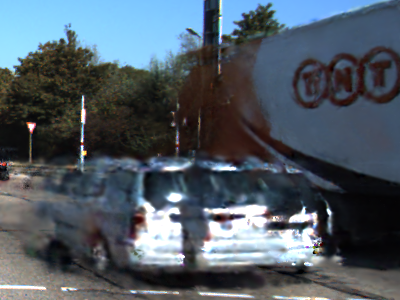}  & 
     \includegraphics[width=\mywidth]{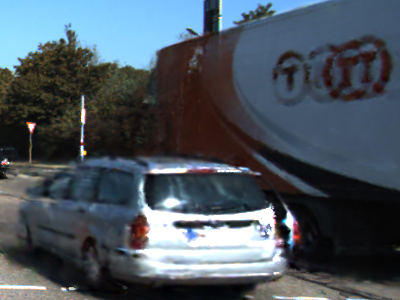} &
     \includegraphics[width=\mywidth]{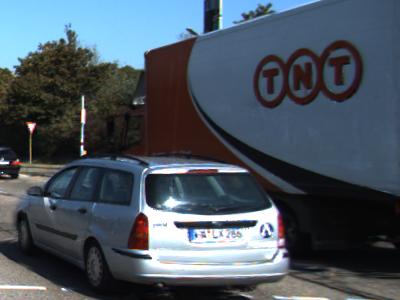} \\
     \includegraphics[width=\mywidth]{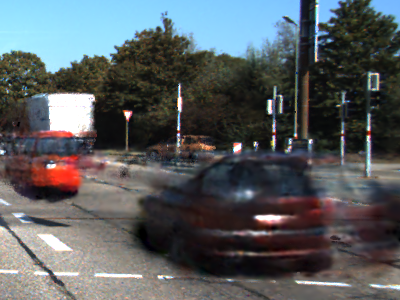}  & 
     \includegraphics[width=\mywidth]{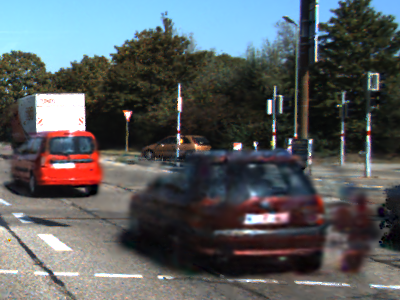} &
     \includegraphics[width=\mywidth]{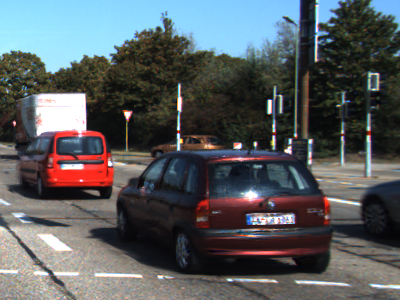} \\
     \rotatebox{0}{w/o opt., w/o uni.} & \rotatebox{0}{w/ opt., w/o uni.} & \rotatebox{0}{Ours} \\
     \end{tabular}
     \vspace{-0.3cm}
     \caption{\textbf{Detail Qualitative Comparison} on KITTI with Noisy Bounding Boxes. }
     \vspace{-0.2cm}
\label{fig:comp_noisy}
\end{figure}

\section{Conclusion}
In this paper, we present HUGS, a holistic scene representation that jointly optimizes appearance, geometry, and motion for urban scenes. This leads to state-of-the-art performance on various tasks. Our method has several limitations. Firstly, the reconstructed dynamic objects can only rotate to a certain degree. Future work may explore category-level prior, to enable accurate reconstruction of the full object. Further, our model lacks control of more degrees of freedom, e.g., light editing, which could be a promising direction to explore based on the Gaussian representation.

\vspace{0.22cm}
{\small
\noindent\textbf{Acknowledgements:}
This work is supported by NSFC under grant 62202418, U21B2004 and the National Key R\&D Program of China under Grant 2021ZD0114501. Yiyi Liao is with the Zhejiang Provincial Key Laboratory of Information Processing, Communication and Networking (IPCAN). Andreas Geiger was supported by the ERC Starting Grant LEGO-3D (850533) and the DFG EXC number 2064/1 - project number 390727645.
}

{
    \small
    \clearpage
    \bibliographystyle{ieeenat_fullname}
    \bibliography{references.bib}
}

\onecolumn
\vspace{5mm}
\appendix
\section*{\Large Appendix}
\renewcommand*{\thesection}{\Alph{section}}
\newcommand{\multiref}[2]{\cref{#1}--\ref{#2}}

In this appendix, we begin by discussing implementation details in \cref{sec:implementation}, which includes information about our 3D Gaussian, metrics, and the training and inference processes. We then describe the datasets used in our experiments in \cref{sec:data}. \cref{sec:baseline} provides information about the baselines we compare with. Finally, \cref{sec:exp} contains additional experiment results.

\section{Implementation}
\label{sec:implementation}
In this section, we begin by discussing our 3D Gaussian details, encompassing semantic, opacity and depth implementation (\cref{sec: 3d_gs}). Subsequently, we discuss the difference between 3D softmax and 2D softmax in 3D Semantic Scene Reconstruction (\cref{sec: 3d_semantic}). Finally, we elucidate the evaluation metrics we utilize (\cref{sec: metrics}). Our source code will be released.
\subsection{3D Gaussian Details}
\label{sec: 3d_gs}

Following \cite{kerbl_3d_2023}, each Gaussian has the following attributes: rotation ($\bR_g \in \nR^{3\times3}$), scale ($\bS_g \in \nR^{3\times1}$), opacity ($\alpha$) and spherical harmonics ($SH$). The corresponding 3D covariance matrix $\bSigma \in \nR^{3\times3}$ can be calculated using the following formula:
\begin{equation}
\bSigma = \bR_g\bS_g\bS_g^T\bR_g^T
\end{equation}
When provided with a viewing transformation $\bW \in \nR^{3\times3}$ and the Jacobian of the affine approximation of the projective transformation $\bJ \in \nR^{3\times3}$, the covariance matrix $\bSigma^{\prime} \in \nR^{3\times3}$ in camera coordinates can be expressed as:
\begin{equation}
\bSigma^{\prime} = \bJ\bW\bSigma \bW^T\bJ^T
\end{equation}
Following EWA splatting \cite{zwicker_ewa_2002}, we can skip the third row and column of $\bSigma^{\prime}$ to obtain a $2\times2$ covariance matrix with the same structure and properties. For brevity, we still use the notation $\bSigma^{\prime} \in \nR^{2 \times 2}$ to denote the 2D covariance matrix.

By considering the projected 3D Gaussian center $\bmu \in \nR^{2\times1}$ and an arbitrary point $\bx \in \nR^{2\times1}$ on camera coordinates, the opacity $\alpha^{\prime}$ of $\bx$ contributed by this 3D Gaussian can be computed as follows:
\begin{equation}
\alpha^{\prime} = \alpha \exp \left( -\frac{1}{2}(\bx-\bmu)^T (\bSigma^{\prime})^{-1} (\bx-\bmu)\right)
\end{equation}
The color $\bc$ of each Gaussian can be computed based on the view direction and its corresponding spherical harmonics ($SH$). Given a set of sorted 3D Gaussians $\cN$ along the ray, we obtain the accumulated color via volume rendering:
\begin{equation}
\pi: \quad \bC = \sum_{i \in \mathcal{N}} \bc_i \alpha'_i \prod_{j=1}^{i-1}(1-\alpha'_j)
\end{equation}
The same volume rendering technique can be applied to obtain semantic $\bS$, depth $\bD$ and optical flow $\bF$. With the given semantic feature $\bs_i$, depth value $d_i$, and Gaussian motion $\mathbf{f}_i$ relative to the camera pose, we can define the semantic rendering, depth rendering, and flow rendering as follows:
\begin{align}
    \label{eq:3d_smt_render}
    \quad \bS &= \sum_{i \in \mathcal{N}} \text{softmax}(\bs_i) \alpha'_i \prod_{j=1}^{i-1}(1-\alpha'_j) \\ 
    \quad \bD &= \sum_{i \in \mathcal{N}} d_i \alpha'_i \prod_{j=1}^{i-1}(1-\alpha'_j)  \\ 
    \quad \bF &= \sum_{i \in \mathcal{N}} \mathbf{f}_i \alpha'_i \prod_{j=1}^{i-1}(1-\alpha'_j)
\end{align}
Note that all the projections and volume rendering techniques mentioned are implemented in CUDA. Calculating the projected 2D opacity $\alpha^{\prime}$ on each pixel and sorting Gaussians based on their distances from the camera takes the majority of computations in the rendering process. These computations need to be performed only once for rendering all modalities, thus maintaining the real-time rendering property of the original 3D Gaussian Splatting. 

\subsection{3D Semantic Scene Reconstruction}
\label{sec: 3d_semantic}
We utilize \cref{eq:3d_smt_render}, referred to as 3D softmax, to render semantic maps. This is in contrast to most existing NeRF-based semantic reconstruction methods that perform softmax to the accumulated 2D logits~\cite{fu_panoptic_2022, zhi_-place_2021}, described in \cref{eq:2d_smt_render}, referred to as 2D softmax. The fundamental difference between these two rendering techniques lies in the fact that 3D softmax normalizes the logits of each 3D point. This normalization process helps prevent a single point with a significantly high logit value from imposing an overwhelming influence on the overall volume rendering outcome. On the other hand, it also prevents placing 3D points of low logit values in empty space. As a result, 3D softmax is effective in reducing floaters and enhancing the geometry of the reconstruction results. In \cref{sec:softmax_ablation}, we present a comprehensive analysis of the qualitative and quantitative comparison results between these two rendering methods.

\begin{equation}
 \quad \bS_{\text{2D\_norm}} = \text{softmax} \left(\sum_{i \in \mathcal{N}} \bs_i \alpha'_i \prod_{j=1}^{i-1}(1-\alpha'_j)\right)
\label{eq:2d_smt_render}
\end{equation}
In the following sections, we refer to our default setting obtained by \cref{eq:3d_smt_render} as $\bS_{\text{3D\_norm}}$.

\subsection{Metrics}
\label{sec: metrics}

\boldparagraph{Novel View Appearance Synthesis}To assess the quality of novel view appearance synthesis, we utilize the Peak Signal-to-Noise Ratio (PSNR), Structural Similarity Index (SSIM), and Learned Perceptual Image Patch Similarity (LPIPS)~\cite{zhang_unreasonable_2018} following the common practice. 

\boldparagraph{Novel View Semantic Synthesis} Following KITTI-360~\cite{liao_kitti-360_2022}, we evaluate the quality of novel view semantic synthesis via the mean Intersection over Union (mIoU) metric.

\boldparagraph{3D Semantic Reconstruction} 
We evaluate 3D semantic reconstruction quality by extracting a 3D semantic point cloud and comparing it with the ground truth LiDAR points. We evaluate both geometric and semantic metrics in the 3D space. Specifically, we evaluate geometric reconstruction quality by measuring the accuracy ($acc.$) and completeness ($comp.$). Accuracy measures the average distance from reconstructed points to the nearest LiDAR point, while completeness measures the average distance from LiDAR points to the nearest reconstructed points. In order to measure the semantic quality of the reconstructed point cloud, we map the predicted 3D semantics to the LiDAR points. Concretely, for each point in the LiDAR point cloud, we identify its closest counterpart in the predicted semantic point cloud and allocate a semantic label based on this nearest neighbor. The assigned semantic labels of all LiDAR points are then compared with the 3D semantic segmentation ground truth provided by KITTI-360, evaluated via the mIoU metric. Note that we only use the LiDAR point clouds for evaluation.

\boldparagraph{3D Tracking} To demonstrate the effectiveness of our model in rectifying noisy 3D tracking results, we evaluate the accuracy of predicted poses compared to ground truth poses in our ablation study. 
Considering the rotation and translation parameters of a ground truth bounding box denoted as $\hat{\bR}$ and $\hat{\bt}$, respectively, and the corresponding parameters of predicted poses, represented as $\bR$ and $\bt$, we employ two metrics for this evaluation following \cite{chen_category_2020}: $e_\bR$ quantifies the rotation accuracy, while $e_\bt$ assesses the translation accuracy as follows
\begin{align}
e_\bR &= \arccos{\frac{Tr(\hat{\bR} \cdot \bR^{-1}) - 1} {2}} \\
e_\bt &= \Vert \hat{\bt} - \bt \Vert_2
\end{align}
where $Tr$ represents the trace of a matrix. 

\boldparagraph{Depth Estimation} In our ablation study, we evaluate the depth estimation quality of our different variants. This is achieved by first projecting the LiDAR points acquired at the same frame to the 2D image space, followed by measuring the L2 distance between the projected LiDAR depth and our method. Considering the projected LiDAR depth is sparse, our assessment focuses solely on pixels with valid LiDAR projections when calculating the L2 distance.

\section{Data}
\label{sec:data}
In this section, we present details of datasets on which we conducted our experiments, including KITTI \cite{geiger_are_2012}, Virtual KITTI 2 (vKITTI) \cite{cabon_virtual_2020} and KITTI-360 \cite{liao_kitti-360_2022}.

\boldparagraph{KITTI}
Following NSG \cite{ost_neural_2021} and MARS \cite{wu_mars_2023}, we select frames 140 to 224 from Scene02 and frames 65 to 120 from Scene06 on KITTI for conducting our experiments.

\boldparagraph{vKITTI}
Virtual KITTI 2 is a synthetic dataset that closely resembles the scenes present in KITTI. In line with the settings outlined in NSG and MARS, we conduct experiments on exactly the same frames from Scene02 and Scene06.

\boldparagraph{KITTI-360}
In addition, we perform experiments on KITTI-360, encompassing both static and dynamic scenes. For the tasks of novel view synthesis and novel semantic synthesis on the leaderboard, we conduct experiments on the sequences provided by the official dataset. Furthermore, we explore dynamic scenes, such as frames 11322 to 11381 from sequence 00, as showcased in our teaser.

\section{Baselines}
\label{sec:baseline}
In this section, we discuss the baselines against which we compare our approach, including  NSG~\cite{ost_neural_2021}, MARS~\cite{wu_mars_2023}, PNF~\cite{kundu_panoptic_2022}, and Semantic Nerfacto~\cite{tancik_nerfstudio_2023}.

\boldparagraph{NSG}
NSG is the pioneering method that introduces the decomposition of dynamic scenes into static background and dynamic foreground components. They propose a learned scene graph representation that enables efficient rendering of novel scene arrangements and viewpoints. However, the official source code provided by NSG often encounters issues when training on KITTI Scene02. Therefore, we utilize the version implemented by the authors of MARS, which is more stable and yields slightly improved results compared to the original version.

\boldparagraph{MARS}
We utilize the latest version of the code provided by the official MARS repository. This latest version incorporates bug fixes and includes additional training iterations, resulting in improved performance. In fact, the updated version achieves a notable improvement of 3 to 4 dB on PSNR compared to the numbers reported in the original paper.

\boldparagraph{PNF}
Since PNF is not open-source, we directly compare our method to their submission on the KITTI-360 leaderboard regarding novel view appearance \& semantic synthesis. To the best of our knowledge, PNF is the only work that considers the optimization of noisy 3D bounding boxes of dynamic objects. In our ablation study, we conduct a na\"ive baseline that optimizes the 3D bounding boxes of each frame independently, which can be considered as a re-implementation of PNF's bounding box optimization in our framework.

\boldparagraph{Semantic Nerfacto}
For the evaluation of 3D semantic point cloud geometry, we compare our results with Semantic Nerfacto \cite{tancik_nerfstudio_2023} as an alternative to PNF \cite{kundu_panoptic_2022}.
Nerfacto \cite{tancik_nerfstudio_2023} is an integration of several successful methods that demonstrate strong performance on real data. It incorporates camera pose refinement, per-image appearance embedding, proposal sampling, scene contraction, and hash encoding within its pipeline. Additionally, Nerfacto includes a semantic head in its framework, enabling the generation of meaningful semantic maps, as demonstrated in \cref{sec:smt_nerfacto}.

\section{Additional Experiment Results}
\label{sec:exp}

\begin{table*}[t]
\centering
\small
{
\setlength{\tabcolsep}{5pt}
\begin{tabular}{lcccccc}
\toprule 
& Pre. & + $\pi$ RGB & + Affine & + $\pi$ Semantic & + $\pi$ Flow  \\
\cline{2-6}
Speed (ms) & 6.25 & 8.13 (+1.88) & 8.54 (+0.41) & 9.70 (+1.16) & 10.17 (+0.47)  \\
\bottomrule
\end{tabular}
}
\caption{\textbf{Time consumption breakdown} of our method.}
\label{tab:breakdown}
\end{table*}

\subsection{Time Consumption Breakdown}
\cref{tab:breakdown} shows our detailed runtime breakdown as various components are incrementally enabled. Preparation (Pre.) contains operations like tile partition and Gaussian sorting. $\pi$ denotes volume rendering, and affine denotes affine transform. Other components like unicycle model, dynamic decomposition, and depth rendering are excluded as they hardly consume any additional time.

\subsection{Additional Comparison Experiments}
\boldparagraph{Dynamic Scene with GT 3D Bounding Boxes}
Despite not being our primary focus, we additionally provide a comparison with NSG and MARS using ground truth 3D trackings. In this setting, our approach demonstrates superior performance across all test scenes, see \cref{tab:comp_gt}.

\begin{table*}[]
\centering
\small
\setlength{\tabcolsep}{4pt}
\begin{tabular}{@{\extracolsep{2pt}}lcccccccccccc@{}} 
\toprule
\multicolumn{1}{c}{} & \multicolumn{3}{c}{KITTI Scene02} & \multicolumn{3}{c}{KITTI Scene06} & \multicolumn{3}{c}{vKITTI Scene02} & \multicolumn{3}{c}{vKITTI Scene06} \\
& PSNR$\uparrow$  & SSIM$\uparrow$  & LPIPS$\downarrow$  & PSNR$\uparrow$  & SSIM$\uparrow$  & LPIPS$\downarrow$ & PSNR$\uparrow$  & SSIM$\uparrow$  & LPIPS$\downarrow$  & PSNR$\uparrow$  & SSIM$\uparrow$  & LPIPS$\downarrow$  \\
\cline{2-4} \cline{5-7} \cline{8-10}  \cline{11-13} 
NSG \cite{ost_neural_2021}  & 22.51 & 0.653 & 0.397 & 23.38 & 0.717 & 0.243 & 23.50 & 0.718 & 0.352 & 26.42 & 0.811 & 0.170 \\
MARS \cite{wu_mars_2023} & 22.95 & 0.728 & 0.145 & 27.01 & 0.883 & 0.062 & 29.80 & 0.950 & 0.034 & 32.71 & 0.959 & 0.023 \\
Ours & \textbf{25.89} & \textbf{0.829} & \textbf{0.092} & \textbf{28.90} & \textbf{0.925} & \textbf{0.016} & \textbf{30.73} & \textbf{0.955} & \textbf{0.018} & \textbf{33.31} & \textbf{0.963} & \textbf{0.010} \\
\bottomrule
\end{tabular}
\caption{\textbf{Novel View Appearance on Dynamic Scenes} with ground truth 3D trackings.}
\label{tab:comp_gt}
\end{table*}

\begin{figure*}[t!]
     \centering
     \small 
     \setlength{\tabcolsep}{0pt}
     \def\mywidth{5.8cm}
     \begin{tabular}{P{\mywidth}P{\mywidth}P{\mywidth}}
 
     \includegraphics[width=\mywidth]{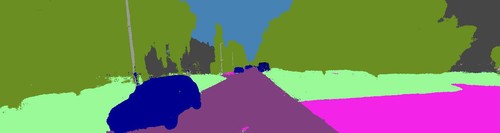}  &
     \includegraphics[width=\mywidth]{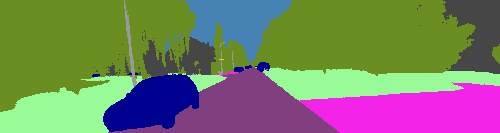}  & \includegraphics[width=\mywidth]{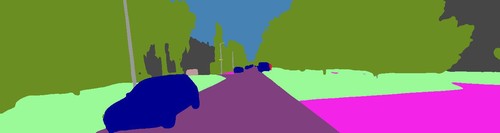} \\

     \includegraphics[width=\mywidth]{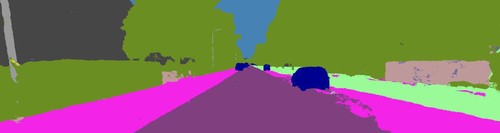}  &
     \includegraphics[width=\mywidth]{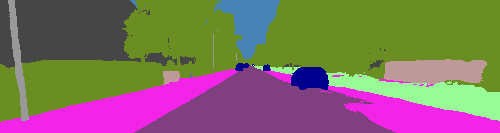}  & \includegraphics[width=\mywidth]{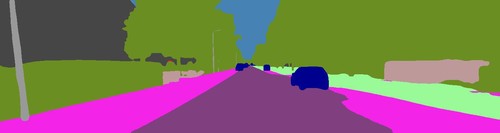} \\
     
     \includegraphics[width=\mywidth]{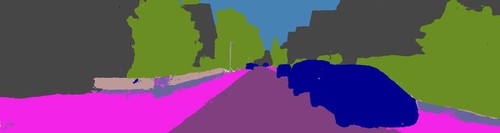}  &
     \includegraphics[width=\mywidth]{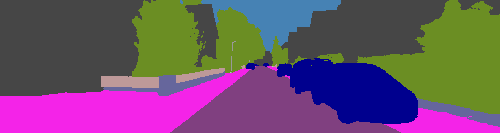}  & \includegraphics[width=\mywidth]{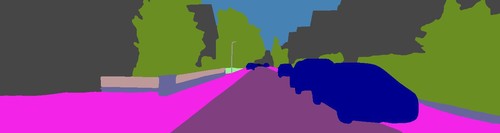} \\

     \includegraphics[width=\mywidth]{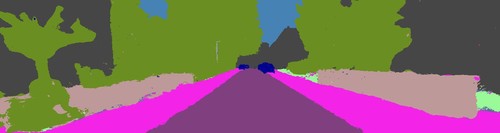}  &
     \includegraphics[width=\mywidth]{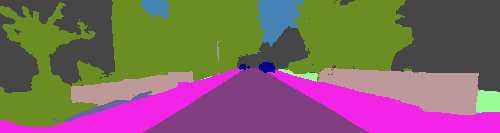}  & \includegraphics[width=\mywidth]{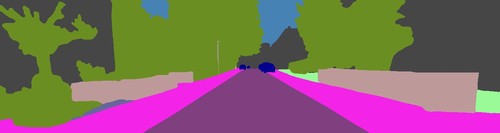} \\
    \rotatebox{0}{Semantic Nerfacto} & \rotatebox{0}{Ours} & \rotatebox{0}{Pseudo GT}
     \end{tabular}
     \caption{\textbf{Qualitative Comparison} with Nerfacto on 2D space. The Pseudo GT column represents the semantic maps that are predicted by \cite{borse_inverseform_2021-1} on GT RGB images.}
\label{fig:compare_nerfacto_2d}
\end{figure*}

\begin{figure}[t!]
     \centering
     \small 
     \setlength{\tabcolsep}{0pt}
     \def\mywidth{7cm}
     \begin{tabular}{P{\mywidth}P{\mywidth}}
 
     \includegraphics[width=\mywidth]{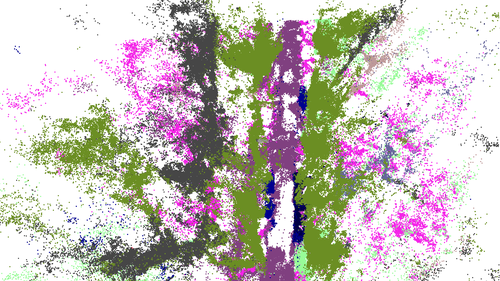}  & \includegraphics[width=\mywidth]{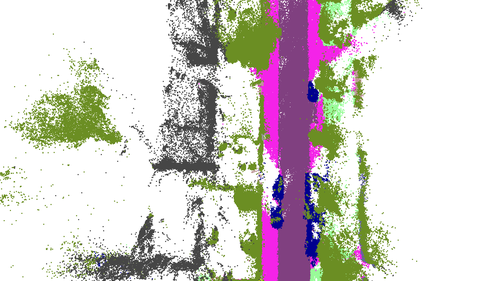} \\

     \includegraphics[width=\mywidth]{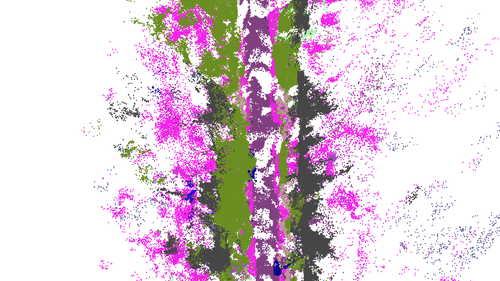}  &
     \includegraphics[width=\mywidth]{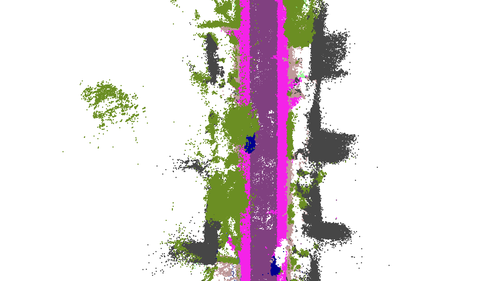} \\
    \rotatebox{0}{Semantic Nerfacto} & \rotatebox{0}{Ours}
     \end{tabular}
     \caption{\textbf{Qualitative Comparison} with Nerfacto on 3D space.
     The semantic point cloud extracted from Semantic Nerfacto struggles to faithfully represent the geometry.
     }
\label{fig:compare_nerfacto_3d}
\end{figure}

\boldparagraph{Details of Comparison with Semantic Nerfacto}
\label{sec:smt_nerfacto}
While Semantic Nerfacto excels at rendering meaningful novel view semantic images (as seen in \cref{fig:compare_nerfacto_2d}), \cref{fig:compare_nerfacto_3d} shows it struggling to accurately reconstruct correct geometry. Following the common practice of NeRF-based semantic reconstruction methods \cite{tancik_nerfstudio_2023}, we apply 2D softmax to Semantic Nerfacto. when we attempted to apply the 3D Softmax technique to Nerfacto, it did not yield better results compared to using 2D softmax. The results can be attributed to the incorrect of Nerfacto's 3D geometry. Instead of adjusting 2D logits with large-scale logits in 3D, the use of 3D softmax prevents the ``cheating'' approach by normalizing logits in 3D space. However, this normalization requirement necessitates sufficiently accurate geometry for satisfactory results.

\begin{figure}[t]
\centering
    \begin{minipage}{0.44\textwidth}
    \centering
    \begin{tabular}{llll}\hline
      \cline{1-4}
                        &       & $e_\mathbf{R}\downarrow$   & $e_\mathbf{t}\downarrow$  \\ 
    \cline{2-4}
    \multirow{2}{*}{KITTI 02} & QD-3DT & 0.027             & 0.215                \\
                        & Ours  & \textbf{0.018}    & \textbf{0.108}       \\ 
    \cline{2-4}
    \multirow{2}{*}{KITTI 06} & QD-3DT & 0.017             & 0.046                \\
                        & Ours  & \textbf{0.012}    & \textbf{0.033}       \\
    \cline{1-4}
    \end{tabular}
    \captionof{table}{\textbf{Qualitative Comparison} with a tracking method, QD-3DT~\cite{hu_monocular_2021}, on two sequences.}
    \label{tab:traj}
    \end{minipage}
    \begin{minipage}{0.44\textwidth}
\centering
\small
\includegraphics[width=0.55\linewidth]{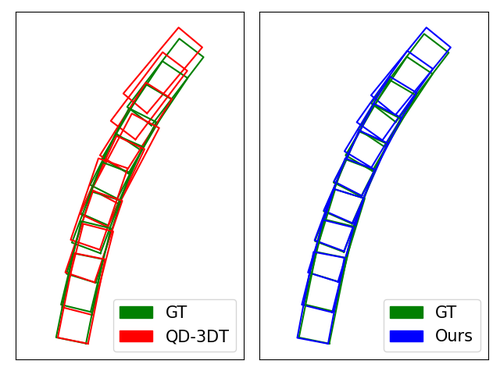} \\
\vspace{-0.3cm}
\captionof{figure}{\textbf{Pose comparison} with QD-3DT.}
\label{fig:traj}
    \end{minipage}
\end{figure}

\boldparagraph{Comparisons with Tracking Methods}
To further compare with off-the-shelf tracking methods, we show the performance of QD-3DT \cite{hu_monocular_2021} and our optimized pose initialized with \cite{hu_monocular_2021} in \cref{tab:traj} and qualitatively illustrate the poses of one vehicle in \cref{fig:traj}. Our method consistently improves \cite{hu_monocular_2021} across two KITTI scenes.

\begin{figure}[t!]
     \centering
     \small 
     \setlength{\tabcolsep}{0pt}
     \def\mywidth{8cm}
     \begin{tabular}{P{\mywidth}P{\mywidth}}
 
     \includegraphics[width=\mywidth]{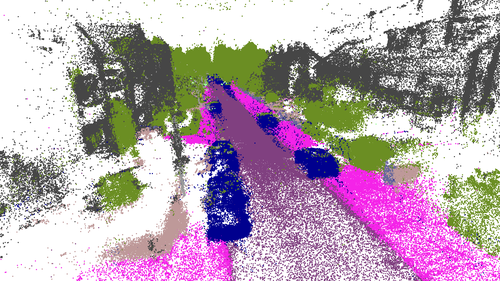}  & \includegraphics[width=\mywidth]{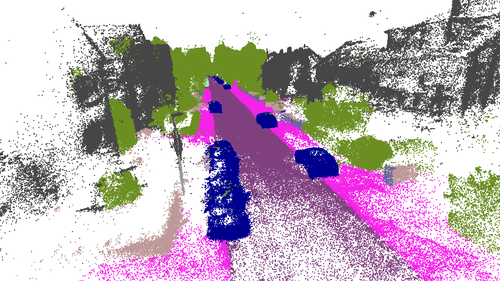} \\

     \includegraphics[width=\mywidth]{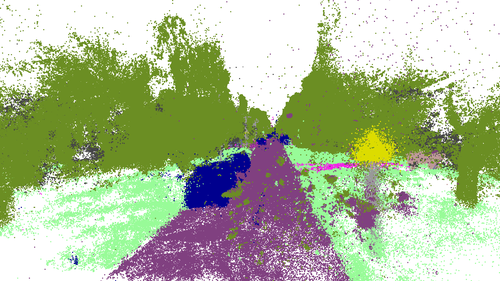}  &
     \includegraphics[width=\mywidth]{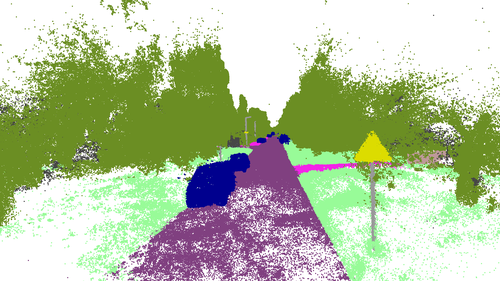} \\

     \includegraphics[width=\mywidth]{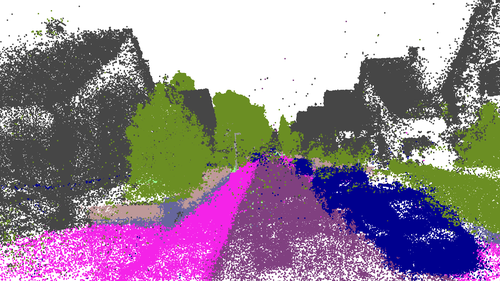}  &
     \includegraphics[width=\mywidth]{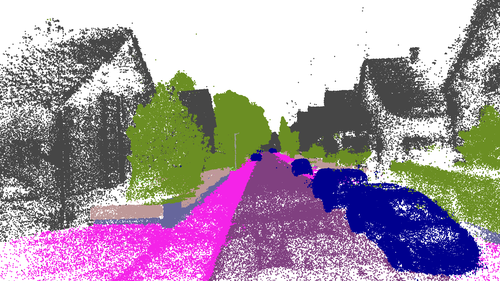} \\
    \rotatebox{0}{Ours w/ $\bS_\text{2D\_norm}$} & \rotatebox{0}{Ours w/ $\bS_\text{3D\_norm}$}
     \end{tabular}
     \caption{\textbf{Qualitative Comparison} of 3D and 2D softmax results. Note that normalizing semantic logits in 3D space (Ours w/ $\bS_\text{3D\_norm}$) clearly reduces floaters and yields better 3D semantic reconstruction than the 2D normalization counterpart (Ours w/ $\bS_\text{2D\_norm}$).}
     \vspace{-0.2cm}
\label{fig:softmax_ablation}
\end{figure}

\subsection{Additional Ablation Experiments}
\boldparagraph{3D and 2D Semantic Softmax}
\label{sec:softmax_ablation}
We provide more 3D and 2D semantic logits softmax comparison in \cref{fig:softmax_ablation} and \cref{tab:semantic2d_3d}. As can be seen, normalizing semantic logits in 3D space leads to notable qualitative and quantitative improvement compared to 2D space normalization.

\begin{table*}[]
\centering
\small
\setlength{\tabcolsep}{14pt}
\begin{tabular}{@{\extracolsep{4pt}}lccc|c@{}} 
\toprule
\multicolumn{1}{c}{} & Seq01 mIoU$_\text{cls}\uparrow$ & Seq02 mIoU$_\text{cls}\uparrow$& Seq03 mIoU$_\text{cls}\uparrow$& Average mIoU$_\text{cls}\uparrow$\\
\hline
Ours w/ $\bS_{\text{2D\_norm}}$ & 0.427 & 0.363 & 0.416 & 0.402 \\
Ours w/ $\bS_{\text{3D\_norm}}$ & \textbf{0.544} &\textbf{0.452} &\textbf{0.520} & \textbf{0.505} \\
\bottomrule
\end{tabular}
\caption{\textbf{Comparison on 3D and 2D Semantic Softmax} on KITTI-360.}
\label{tab:semantic2d_3d}
\end{table*}

\begin{table*}[]
\centering
\small
\setlength{\tabcolsep}{2.1pt}
\begin{tabular}{@{\extracolsep{2pt}}lcccccccccccc@{}} 
\toprule
\multicolumn{1}{c}{} & \multicolumn{3}{c}{KITTI-360 Scene00} & \multicolumn{3}{c}{KITTI-360 Scene01} & \multicolumn{3}{c}{KITTI-360 Scene02} & \multicolumn{3}{c}{Average} \\
& PSNR$\uparrow$  & SSIM$\uparrow$  & LPIPS$\downarrow$  & PSNR$\uparrow$  & SSIM$\uparrow$  & LPIPS$\downarrow$ & PSNR$\uparrow$  & SSIM$\uparrow$  & LPIPS$\downarrow$  & PSNR$\uparrow$  & SSIM$\uparrow$  & LPIPS$\downarrow$  \\
\cline{2-4} \cline{5-7} \cline{8-10}  \cline{11-13} 
Random
& 20.84 & 0.784 & 0.150 & 19.40 & 0.705 & 0.171 & 22.55 & 0.800 & 0.136 & 20.93 & 0.763 & 0.457 \\
LiDAR
& 25.64 & 0.856 & 0.070 & 22.88 & 0.784 & \textbf{0.089} & 24.04 & 0.836 & 0.080 & 24.19 & 0.825 & \textbf{0.080} \\
COLMAP
& \textbf{26.23} & \textbf{0.863} & \textbf{0.069} & \textbf{22.94} & \textbf{0.794} & 0.096 & \textbf{24.38} & \textbf{0.843} & \textbf{0.077} & \textbf{24.52} & \textbf{0.833} & 0.081 \\
\bottomrule
\end{tabular}
\caption{\textbf{Quantitative Comparison} with different initialization.}
\label{tab:init}
\end{table*}

\begin{figure*}[t!]
     \centering
     \small 
     \setlength{\tabcolsep}{0pt}
     \def\mywidth{4.35cm}
     \begin{tabular}{P{\mywidth}P{\mywidth}P{\mywidth}P{\mywidth}}
 
     \includegraphics[width=\mywidth]{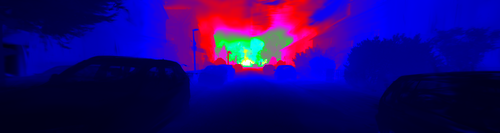}  & \includegraphics[width=\mywidth]{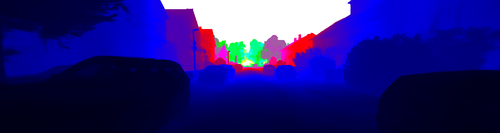}  &
     \includegraphics[width=\mywidth]{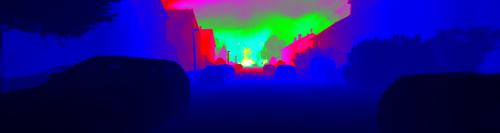}  & \includegraphics[width=\mywidth]{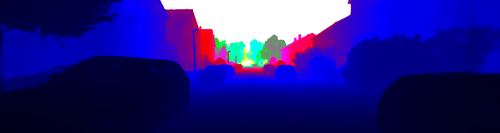}  \\

     \includegraphics[width=\mywidth]{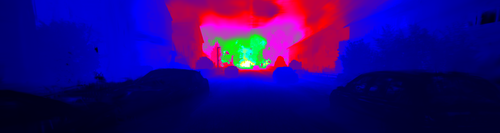}  & \includegraphics[width=\mywidth]{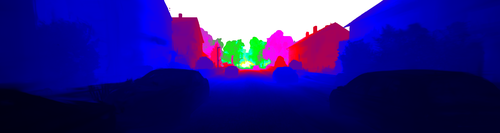}  &
     \includegraphics[width=\mywidth]{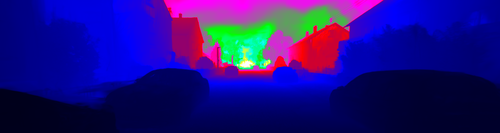}  & \includegraphics[width=\mywidth]{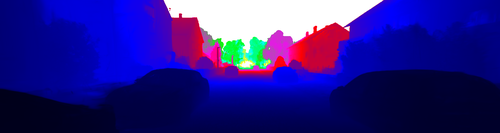}  \\

     \includegraphics[width=\mywidth]{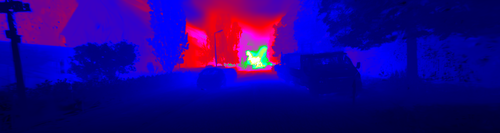}  & \includegraphics[width=\mywidth]{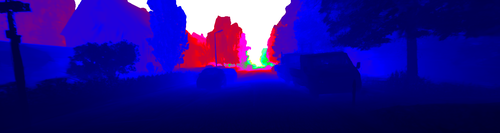}  &
     \includegraphics[width=\mywidth]{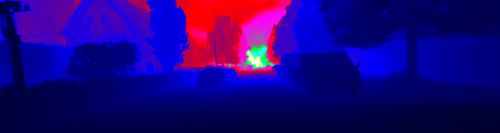}  & \includegraphics[width=\mywidth]{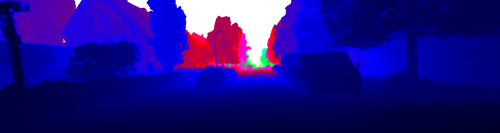}  \\

     \includegraphics[width=\mywidth]{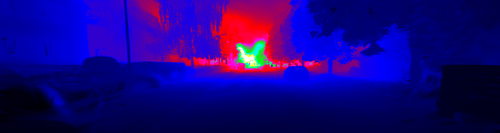}  & \includegraphics[width=\mywidth]{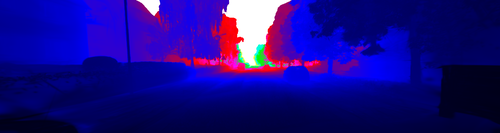}  &
     \includegraphics[width=\mywidth]{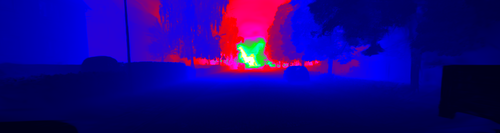}  & \includegraphics[width=\mywidth]{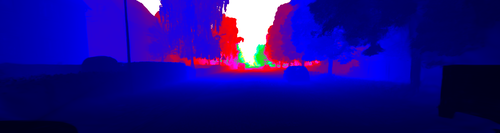}  \\

     \includegraphics[width=\mywidth]{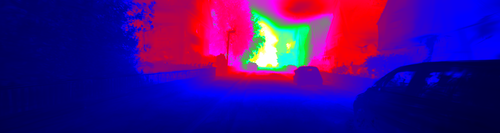}  & \includegraphics[width=\mywidth]{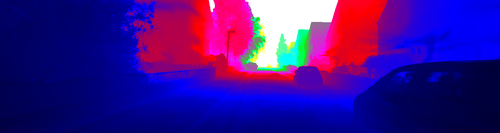}  &
     \includegraphics[width=\mywidth]{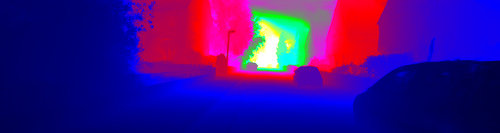}  & \includegraphics[width=\mywidth]{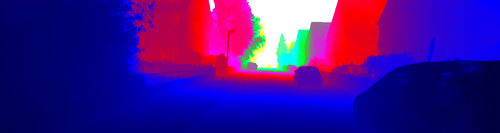}  \\

     \includegraphics[width=\mywidth]{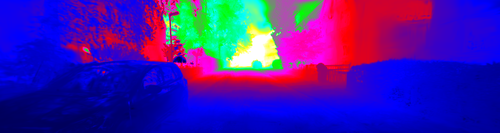}  & \includegraphics[width=\mywidth]{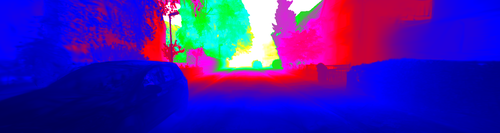}  &
     \includegraphics[width=\mywidth]{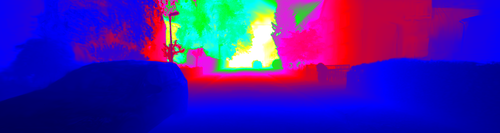}  & \includegraphics[width=\mywidth]{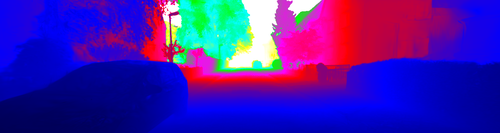}  \\

    \rotatebox{0}{w/o $\cL_\bS$, w/o  $\cL_\bF$} & \rotatebox{0}{w/ $\cL_\bS$, w/o  $\cL_\bF$}  & \rotatebox{0}{w/o  $\cL_\bS$, w/  $\cL_\bF$}  & \rotatebox{0}{w/  $\cL_\bS$, w/ $\cL_\bF$}
     \end{tabular}
     \caption{\textbf{Qualitative Comparison} on depth. In the presence of the semantic loss $\cL_\bS$ (2nd and 4th columns), we set the sky region's depth infinite based on its semantic label.  Note that the activation of either the semantic loss $\cL_\bS$ (2nd column) or the optical loss $\cL_\bF$ (3rd column) yields enhancements in geometry, e.g., the left car in the bottom row, with the improvement in optical flow loss being more evident.}
\label{fig:depth_comp}
\end{figure*}

\begin{figure*}[t!]
     \centering
     \small 
     \setlength{\tabcolsep}{0pt}
     \def\mywidth{4.35cm}
     \begin{tabular}{P{\mywidth}P{\mywidth}P{\mywidth}P{\mywidth}}
 
     \includegraphics[width=\mywidth]{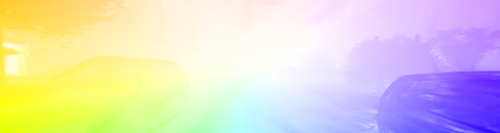}  & \includegraphics[width=\mywidth]{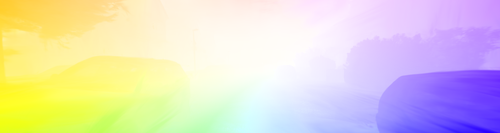}  &
     \includegraphics[width=\mywidth]{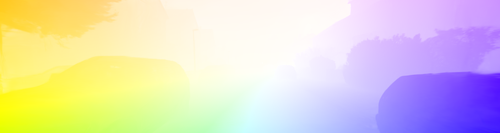}  & \includegraphics[width=\mywidth]{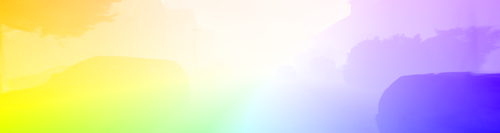}  \\

     \includegraphics[width=\mywidth]{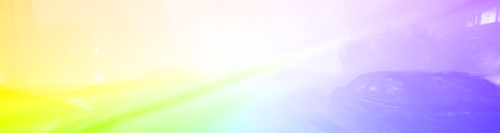}  & \includegraphics[width=\mywidth]{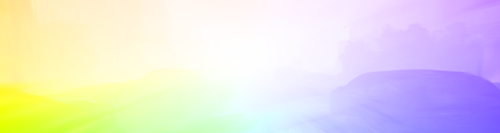}  &
     \includegraphics[width=\mywidth]{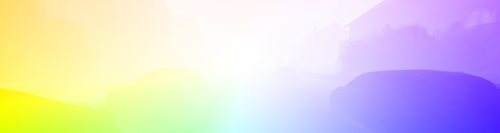}  & \includegraphics[width=\mywidth]{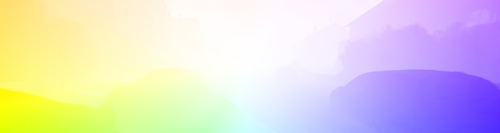}  \\

     \includegraphics[width=\mywidth]{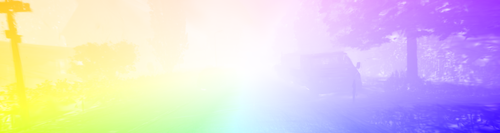}  & \includegraphics[width=\mywidth]{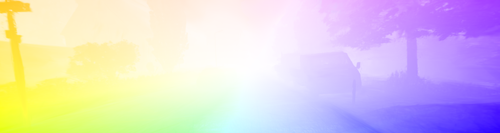}  &
     \includegraphics[width=\mywidth]{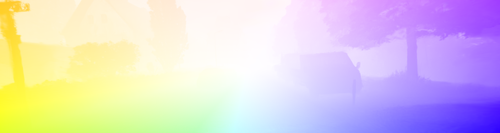}  & \includegraphics[width=\mywidth]{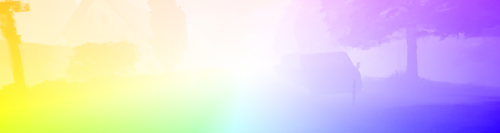}  \\

     \includegraphics[width=\mywidth]{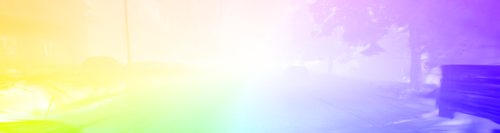}  & \includegraphics[width=\mywidth]{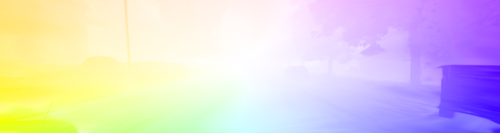}  &
     \includegraphics[width=\mywidth]{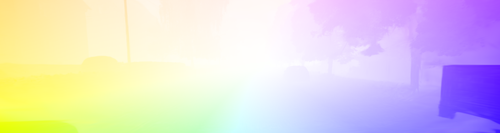}  & \includegraphics[width=\mywidth]{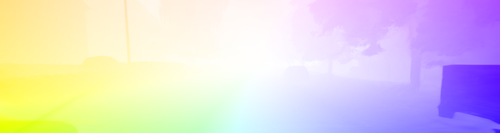}  \\

     \includegraphics[width=\mywidth]{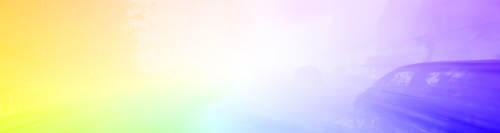}  & \includegraphics[width=\mywidth]{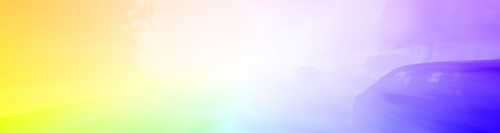}  &
     \includegraphics[width=\mywidth]{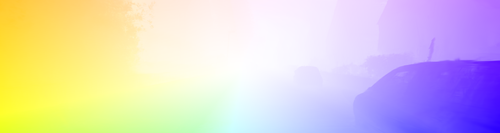}  & \includegraphics[width=\mywidth]{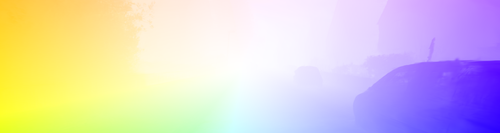}  \\

     \includegraphics[width=\mywidth]{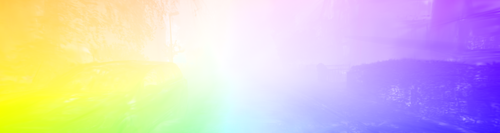}  & \includegraphics[width=\mywidth]{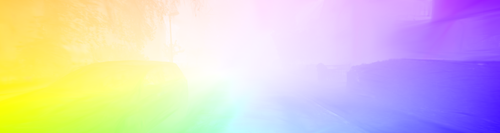}  &
     \includegraphics[width=\mywidth]{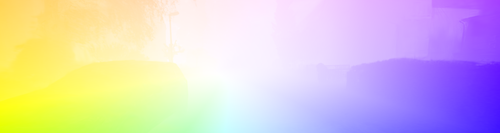}  & \includegraphics[width=\mywidth]{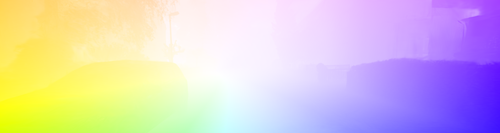}  \\

    \rotatebox{0}{w/o $\cL_\bS$, w/o  $\cL_\bF$} & \rotatebox{0}{w/ $\cL_\bS$, w/o  $\cL_\bF$}  & \rotatebox{0}{w/o  $\cL_\bS$, w/  $\cL_\bF$}  & \rotatebox{0}{w/  $\cL_\bS$, w/ $\cL_\bF$}
     \end{tabular}
     \caption{\textbf{Qualitative Comparison} on optical flow. While 3D Gaussians can enable the rendering of optical flow without additional supervision on semantic or optical flow, the rendered flow maps exhibit clear artifacts (1st column). These artifacts are particularly noticeable on the cars and the ground. Interestingly, the incorporation of semantic supervision $\cL_{\bS}$ mitigates the artifacts to some extent (2nd column). Additionally, introducing pseudo-optical flow supervision $\cL_{\bF}$ contributes to further improvement in the optical flow results (3rd and 4th columns).}
\label{fig:flow_comp}
\end{figure*}

\boldparagraph{Improvements on Geometry}
We now qualitatively examine how the optical flow loss $\cL_\bF$ and the semantic loss $\cL_\bS$ impact the geometry, as shown in \cref{fig:depth_comp} and \cref{fig:flow_comp}. Both figures reveal that incorporating either the semantic loss or the optical flow loss improves the underlying geometry. While the impact of the semantic loss on geometry may be less evident, the optical flow clearly enhances geometric accuracy. This improvement is rationalized by the fact that optical flow guides correspondences across neighboring frames. It's important to note that when the semantic loss $\cL_\bS$ is active, the sky region of the depth maps in \cref{fig:depth_comp} is set to infinite.

\begin{figure*}[t!]
     \centering
     \small 
     \setlength{\tabcolsep}{0pt}
     \def\mywidth{5.8cm}
     \begin{tabular}{P{\mywidth}P{\mywidth}P{\mywidth}}
 
     \includegraphics[width=\mywidth]{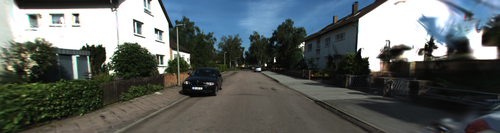}  & \includegraphics[width=\mywidth]{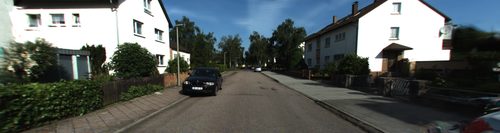}  &
     \includegraphics[width=\mywidth]{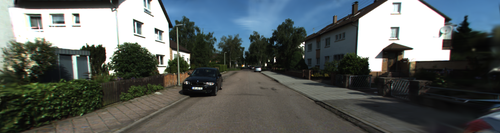}  \\

    \includegraphics[width=\mywidth]{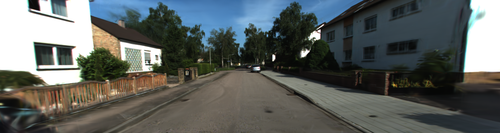}  & \includegraphics[width=\mywidth]{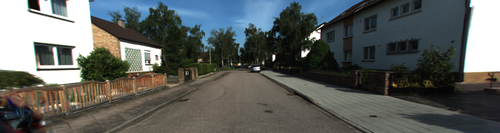}  &
    \includegraphics[width=\mywidth]{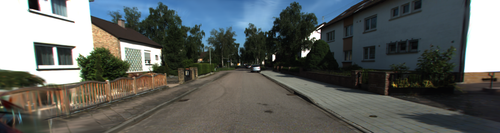}  \\

    \includegraphics[width=\mywidth]{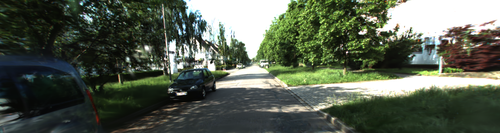}  & \includegraphics[width=\mywidth]{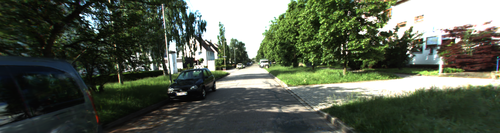}  &
    \includegraphics[width=\mywidth]{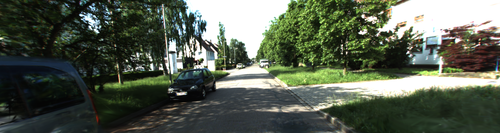}  \\

    \includegraphics[width=\mywidth]{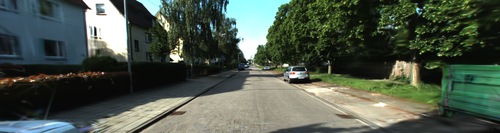}  & \includegraphics[width=\mywidth]{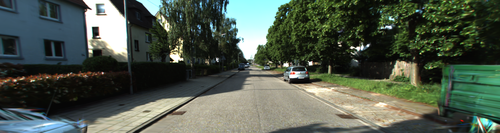}  &
    \includegraphics[width=\mywidth]{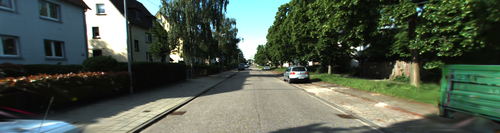}  \\

    \includegraphics[width=\mywidth]{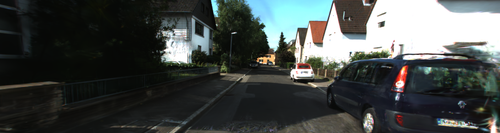}  & \includegraphics[width=\mywidth]{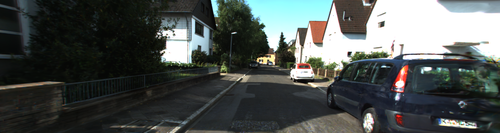}  &
    \includegraphics[width=\mywidth]{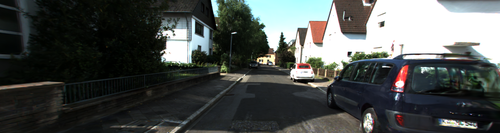}  \\

    \includegraphics[width=\mywidth]{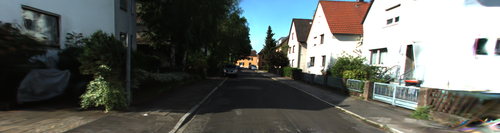}  & \includegraphics[width=\mywidth]{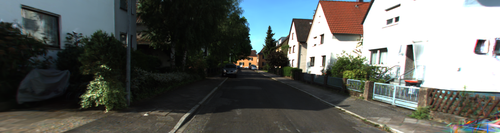}  &
    \includegraphics[width=\mywidth]{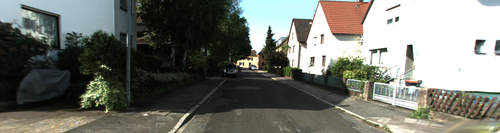}  \\

    \rotatebox{0}{Random} & \rotatebox{0}{LiDAR}  & \rotatebox{0}{COLMAP}
     \end{tabular}
     \caption{\textbf{Qualitative Comparison} with different initialization strategies. The superiority of both LiDAR-based and COLMAP-based initialization over random initialization is evident. Random initialization occasionally results in significant artifacts, as illustrated by the right building in the 1st row. LiDAR-based initialization, while generally effective, introduces artifacts in areas very close to the ego car, such as the bottom right corner of the 4th-6th rows. These regions typically encompass LiDAR points unseen by any training views. The COLMAP-based initialization further demonstrates an improvement over the LiDAR-based approach in distant regions, exemplified by the trees in the 1st row. }
     \vspace{-0.5cm}
\label{fig:init}
\end{figure*}

\boldparagraph{Effects of Initialization}
We conduct a thorough comparison of the results obtained through different initialization strategies. In particular, we consider random initialization and COLMAP-based initialization. To further investigate whether adopting LiDAR point cloud for initialization is helpful in urban scenes, we further consider LiDAR point clouds as initialization. We report the quantitative and qualitative comparison in \cref{tab:init} and \cref{fig:init}, respectively.
We observe that both LiDAR and COLMAP initialization outperform random initialization. Interestingly, the COLMAP-based initialization even shows a slight advantage over the LiDAR-based one. This could be attributed to the presence of points in the LiDAR point clouds that remain unobserved in any training views, leading to artifacts in test viewpoints. Furthermore, COLMAP improves the quality of objects located at far distances, which cannot be accurately captured by LiDAR.  These findings underscore the potential for achieving high-fidelity novel view synthesis in urban scenes based solely on RGB images. In our main experiments, we adopt the COLMAP-based initialization by default.

\subsection{Visualization of Optimization Progress}
\begin{figure*}[t!]
     \centering
     \small 
     \setlength{\tabcolsep}{0pt}
     \def\mywidth{6cm}
     \begin{tabular}{P{\mywidth}P{\mywidth}P{\mywidth}}
 
     \includegraphics[width=\mywidth]{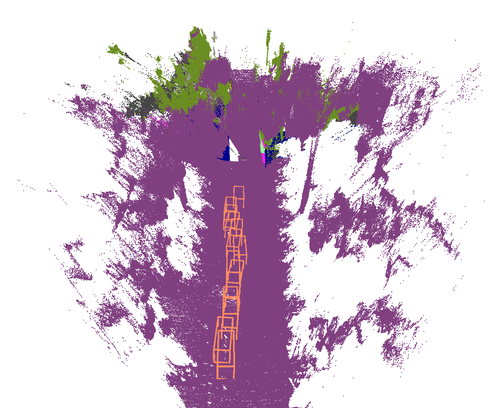}  & 
     \includegraphics[width=\mywidth]{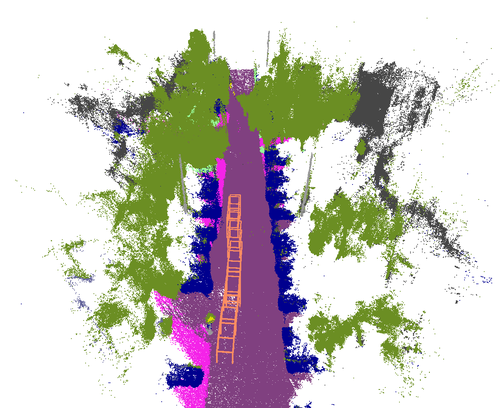}  &
     \includegraphics[width=\mywidth]{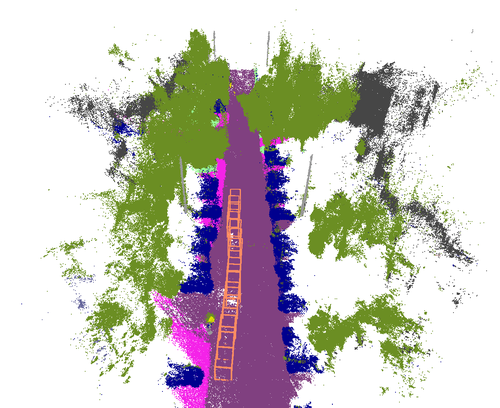} \\

    \rotatebox{0}{10 steps} & \rotatebox{0}{2000 steps} & \rotatebox{0}{5000 steps}
     \end{tabular}
    \vspace{-0.3cm}
     \caption{\textbf{Visualization of Optimization Progress}. Our method jointly optimizes the static background and the trajectory of the dynamic foreground objects. By integrating physical constraints using the unicycle model, our method allows for recovering a smooth trajectory from noisy 3D bounding boxes. To prevent visual clutter, we exclude point clouds of the dynamic object and only visualize the bounding boxes.}
     \vspace{-0.2cm}
\label{fig:vis_optim}
\end{figure*}

We present the visualization of the optimization progress for both the noisy bounding boxes and the background semantic point cloud in \cref{fig:vis_optim}. 
Using noisy 3D bounding boxes as input, our approach optimizes both the background and the poses of the bounding boxes simultaneously. As evident, the application of physical constraints derived from the unicycle model results in a smooth trajectory for the bounding boxes.

\end{document}